\documentclass[runningheads]{llncs}

\usepackage{eccv}

\usepackage{eccvabbrv}

\usepackage{graphicx}
\usepackage{booktabs}

\usepackage[accsupp]{axessibility}  

\usepackage{hyperref}

\usepackage{orcidlink}

\usepackage[table]{xcolor}
\usepackage{siunitx}
\usepackage{rotating}   
\usepackage{makecell}   
\usepackage{multirow}
\usepackage{amssymb} 
\usepackage{pifont} 
\usepackage{xcolor} 

\definecolor{myblue}{RGB}{230, 240, 255}
\definecolor{tablegray}{gray}{0.92}

\usepackage{bbding}
\usepackage{pdfpages}

\begin{document}

\title{SpaR3D-MoE: Adaptive 3D Spatial Reasoning from Sparse Views Meets Geometry-Inductive Mixture-of-Experts} 

\titlerunning{SpaR3D-MoE}

\author{Haida Feng\inst{1,2}\orcidlink{0000-0002-5862-3134} \and
Hao Wei\inst{1,3}\textsuperscript{\textdagger}\orcidlink{0000-0002-7708-0243} \and
Haolin Wang\inst{1,2}\orcidlink{0009-0008-8073-9492} \and Shiwei Li\inst{1,2}\orcidlink{0009-0006-1973-6810} \and Chade Li\inst{1,2}\orcidlink{0009-0002-5558-4739} \and Yihong Wu\inst{1,2,3}\textsuperscript{\textdagger}\orcidlink{0000-0003-2198-9113}}

\authorrunning{Feng et al.}

\institute{State Key Laboratory of Multimodal Artificial Intelligence Systems,
Institute of Automation, Chinese Academy of Sciences, China \and
School of Artificial Intelligence, University of Chinese Academy of Sciences, China \and YUKUN Intelligent World, Beijing, China\\
\email{\{fenghaida2024@ia, weihao2019@ia, wanghaolin2023@ia, lishiwei2023@ia, lichade2021@ia, yhwu@nlpr.ia\}.ac.cn}}

\maketitle
\begingroup
\renewcommand{\thefootnote}{\textdagger}
\footnotetext{Corresponding authors}
\endgroup

\begin{abstract}
Recent Multimodal Large Language Models (MLLMs) struggle to bridge the representational gap between 2D semantic understanding and 3D spatial geometry. Existing 3D-aware models either rely on costly 3D-specific data or utilize RGB-only inputs with heuristic sampling and monolithic, shallow fusion, which respectively disrupt essential spatiotemporal connectivity and induce modality contention across diverse spatial tasks. To overcome these bottlenecks, we introduce SpaR3D-MoE, an end-to-end framework that enables adaptive spatial reasoning by equipping MLLMs with geometry-aware capabilities from only sparse RGB inputs. First, we propose an adaptive spatiotemporal manifold sampling mechanism  that constructs a geometry-aware spatiotemporal graph to extract informative keyframes, effectively mitigating sequence redundancy while preserving the scene's topological connectivity. Second, we introduce the heterogeneous geometry-inductive Mixture-of-Experts driven by an instruction-pose aware router, which adaptively routes multimodal tokens to specialized experts, resolving the cross-modal contention inherent in monolithic fusion. Extensive experiments on VSI-Bench, ScanQA, and SQA3D demonstrate that our method achieves state-of-the-art performance. Notably, SpaR3D-MoE achieves the highest average score of 63.5 on VSI-Bench, outperforming the strongest baseline by 7.8 absolute points, alongside relative improvements of 35.4\% and 51.4\% in Route Plan and Relative Direction tasks, respectively.

\end{abstract}

\section{Introduction}
\begin{figure}[t]
    \centering
    \includegraphics[width=\linewidth]{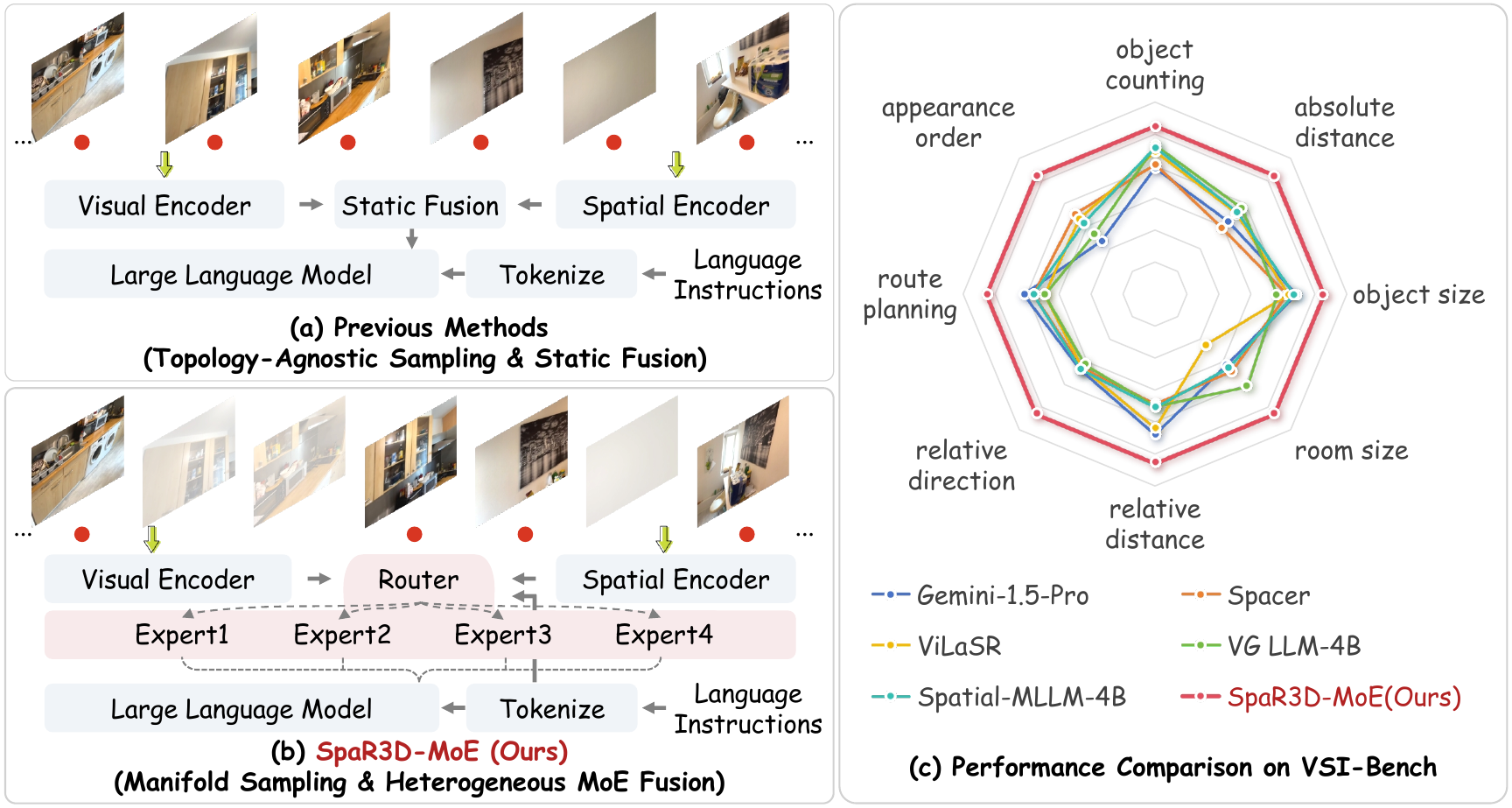}

    
    \caption{\textbf{Comparison of 3D spatial reasoning paradigms.} Existing methods use topology-agnostic sampling and static monolithic fusion \textbf{(a)}. SpaR3D-MoE selects sparse keyframes via manifold-based sampling and routes multimodal tokens to specialized experts \textbf{(b)}, achieving strong performance across VSI-Bench spatial tasks \textbf{(c)}.}

    \label{fig:teaser}
\end{figure}

Three-dimensional (3D) spatial reasoning is a cornerstone of Embodied AI, enabling agents to understand complex environments, reason about spatial relationships, and ground natural language instructions in real-world contexts~\cite{anderson2018vision,zhi2025lscenellm}. While Multimodal Large Language Models (MLLMs)~\cite{alayrac2022flamingo,chen2024internvl,liu2023visual,geminiteam2024gemini15unlockingmultimodal,Bai2025Qwen25VLTR,Bai2025Qwen3VLTR,openai2024gpt4ocard} excel at 2D image and video understanding, extending them to the 3D physical world is hindered by a fundamental representational gap. Current models struggle with complex tasks like estimating metric distances or navigating spatial gaps, primarily because bridging 2D visual semantics with geometric 3D real-world alignments remains a challenge.

To bridge this representational gap, recent efforts have diverged into two paradigms aiming to map 3D geometric features into the MLLM latent space, either through explicit 3D structures or implicit visual representations. The first paradigm explicitly integrates 3D structures (e.g., point clouds, depth maps, or reconstructed meshes) directly into Large Language Models (LLMs)~\cite{zhu2024llava-3d, zhi2025lscenellm, wang2023chat-3d, chen2024ll3da, huang2024LEO, huang2024chatscene, fu2024scene-llm, hong20233d, chen2024grounded-3d-llm}. These approaches typically lift multi-view RGB-D inputs or reconstructed scenes into 3D point clouds~\cite{hong20233d}, then employ specialized 3D geometry encoders (e.g., PointNet++~\cite{qi2017pointnet++}) to extract geometric features and project them into the textual latent space. While these explicit geometric priors significantly benefit physical grounding, this pipeline is fundamentally constrained by its rigid dependence on acquiring explicit 3D geometric proxies. Moreover, the intrinsic sparsity of point clouds often leads to the loss of rich visual details, compromising the fine-grained semantic understanding crucial for comprehensive scene reasoning. Conversely, the second paradigm focuses on scalable, 3D-aware MLLMs that rely solely on RGB sequences~\cite{Zheng2025LearningFV, Wu2025SpatialMLLMBM}. These methods typically adopt a dual-encoder architecture, using a 2D visual encoder to extract semantic features and a spatial encoder, often initialized from visual geometry foundation models~\cite{wang2025vggt}, to recover implicit 3D structural features from 2D RGB inputs. Despite bypassing costly 3D-specific data for easier applicability, this paradigm remains constrained by heuristic sampling and static, monolithic fusion. Regarding frame sampling, current methods rely on topology-agnostic selection strategies, such as rigid uniform sampling or discrete voxel maximization heuristics. By treating frames as isolated viewpoints, these approaches introduce spatiotemporal redundancy and overlook the intrinsic spatiotemporal manifold of the 3D scene, potentially missing critical spatial frames. At the multimodal integration level, monolithic fusion indiscriminately projects heterogeneous semantic textures and geometric structures into a shared latent space. Such architectural inflexibility induces cross-modal contention when faced with diverse task requirements, thereby hindering fine-grained spatial reasoning, as illustrated in \cref{fig:teaser}(a).

To address these limitations, we introduce SpaR3D-MoE, a framework that shifts the paradigm toward adaptive spatial reasoning from sparse RGB inputs, as illustrated in \cref{fig:teaser}(b). Specifically, we propose an \textbf{A}daptive \textbf{S}patiotemporal \textbf{M}anifold \textbf{S}ampling (ASMS) mechanism to extract informative keyframes. By constructing a spatiotemporal graph from viewpoint-dependent geometry and camera ego-motion, regulated by a motion-aware quality gate, it filters redundancy while preserving essential spatiotemporal connectivity of the scene. Furthermore, to mitigate the cross-modal contention in monolithic fusion, we present a \textbf{H}eterogeneous \textbf{G}eometry-\textbf{I}nductive Mixture-of-Experts (HGI-MoE). To the best of our knowledge, this is the first work that introduces MoE architectures into 3D spatial reasoning. Inspired by the functional specialization of the human brain~\cite{ungerleider1994and}, this module is driven by an \textbf{I}nstruction-\textbf{P}ose \textbf{A}ware \textbf{R}outer (IPAR), which adaptively routes multimodal features to specialized experts based on linguistic intent and camera ego-motion. Crucially, these experts exhibit emergent specialization, performing varying levels of cross-modal fusion tailored to meet diverse spatial reasoning requirements. This dynamic dispatching effectively mitigates cross-modal interference by establishing disentangled reasoning pathways. To ensure stable expert convergence, we also introduce a load-balancing loss function that regularizes the expert assignment and prevents routing collapse. Extensive experiments on large-scale benchmarks, including VSI-Bench~\cite{yang2025thinking}, ScanQA~\cite{azuma2022scanqa}, and SQA3D~\cite{ma2023sqa3dsituatedquestionanswering}, demonstrate that SpaR3D-MoE achieves state-of-the-art (SOTA) performance, validating its adaptive reasoning across diverse challenging 3D spatial reasoning tasks from sparse RGB-only views.

The main contributions of SpaR3D-MoE are  outlined as follows:
\begin{itemize}
    \item We propose ASMS mechanism, which constructs a spatiotemporal graph with a motion-aware quality gate to adaptively extract informative keyframes, reducing redundancy while preserving manifold connectivity, achieves a 10.6\% improvement over uniform sampling on the VSI-Bench Route Plan task.
    \item We propose HGI-MoE, which introduces the Mixture-of-Experts paradigm to 3D spatial reasoning, where an IPAR adaptively dispatches multimodal features to specialized experts for varying levels of fusion, mitigating modality contention and robustly fulfilling diverse spatial reasoning demands.
    \item Empowered by these designs, we introduce SpaR3D-MoE, an end-to-end adaptive framework that endows MLLMs with physically-grounded spatial intelligence from sparse RGB-only views. It achieves SOTA across diverse benchmarks, notably surpassing the strongest VSI-Bench baseline by 7.8 absolute points, while delivering remarkable relative gains of 35.4\% and 51.4\% in complex navigation and fine-grained metric estimation, respectively.
\end{itemize}

\section{Related Work}
\subsection{Multimodal Large Language Models}
Recent Multimodal Large Language Models (MLLMs) \cite{Bai2025Qwen25VLTR,openai2024gpt4ocard,zhang2024llavanextvideo,geminiteam2024gemini15unlockingmultimodal,Tong2024Cambrian1AF,Bai2025Qwen3VLTR} have achieved impressive capabilities in image and video understanding. These models typically integrate visual encoders with large language models to process and generate text based on visual inputs. They have been applied in various domains, including visual question answering and multimodal dialogue systems. However, while current MLLMs can reason about complex relationships from images and videos, they struggle to ground these relations in the 3D physical world. Primarily trained on 2D image-text pairs that prioritize semantic content over geometric structure, these models lack a necessary understanding of 3D spatial relationships and geometry. Consequently, with standard visual encoders compressing spatial details for semantic alignment \cite{chen2024spatialvlm,Tang2025LEGOPuzzlesHG}, 2D-based MLLMs often hallucinate when faced with complex 3D spatial understanding tasks.

\subsection{3D-Aware Multimodal Large Language Models}
Recent research increasingly leverages pre-trained MLLMs to tackle complex 3D scene understanding and reasoning tasks. Existing methods can be broadly grouped by how they introduce spatial information. Early works~\cite{hong20233d,wang2023chat-3d,huang2024chatscene,chen2024ll3da,fu2024scene-llm,zhu2024llava-3d,chen2024grounded-3d-llm,huang2024LEO,zhi2025lscenellm} mainly rely on 2.5D or 3D representations, such as posed RGB-D data, point clouds, or voxel grids. Although these explicit geometric inputs provide strong 3D grounding, their dependence on depth sensors or reconstructed 3D assets limits scalability in RGB-only scenarios. Recent RGB-based methods~\cite{Zheng2025LearningFV,Wu2025SpatialMLLMBM} alleviate this issue by using VGGT~\cite{wang2025vggt} to extract implicit 3D geometric features from monocular inputs and align them with MLLMs. Other studies improve spatial reasoning from complementary perspectives, such as reinforcement learning for video spatial reasoning~\cite{ouyang2025spacerreinforcingmllmsvideo} and structured 2D representations for perception-guided reasoning~\cite{zhu2025struct2d}. Despite these advances, preserving sparse-view spatiotemporal topology while adaptively fusing visual and geometric features for diverse spatial tasks remains insufficiently explored. In contrast, our ASMS adaptively samples sparse keyframes to preserve essential spatiotemporal connectivity, while HGI-MoE dynamically activates specialized experts for task-aware visual-geometric fusion, thereby enhancing 3D spatial reasoning performance.

\subsection{Mixture-of-Experts Framework}
The MoE paradigm has fundamentally transformed the scaling of LLMs by decoupling model capacity from inference latency \cite{Shen2023FlanMoESI, xue2024openmoe, jiang2024mixtral}. By dynamically routing tokens to a sparse subset of active parameters, these architectures achieve massive representational power without imposing prohibitive computational bottlenecks. Building on this foundation, recent foundation models~\cite{yang2024qwen2technicalreport,qwen2025qwen25technicalreport,yang2025qwen3technicalreport,Dai2024DeepSeekMoETU} have further refined sparse routing mechanisms to achieve unprecedented scale and efficiency in pure language modeling. Beyond text, this paradigm has been extended to the multimodal domain~\cite{lin2026moe,li2025uni}, where modality-specific encoders and multi-stage tuning successfully scale diverse inputs without sparsity degradation. Motivated by the efficacy of MoE in modeling heterogeneous data, we introduce a multimodal MoE framework tailored for various complex 3D spatial reasoning tasks. To accommodate the diverse distributions of data inputs and instructional intents, our framework dynamically routes features to heterogeneous experts, enabling efficient and robust scene comprehension.

\section{Methodology}
\subsection{Problem Formulation and Framework}
Let $\mathcal{V} = \{v_i\}_{i=1}^{N_k}$ denote a continuous sequence of embodied visual observations capturing a complex 3D environment, and $\mathcal{Q}$ represent a natural language instruction specifying a spatial reasoning task. Our primary objective is to autoregressively decode a precise textual response $\mathcal{A} = \{a_t\}_{t=1}^T$.

Processing RGB video sequences for spatial understanding through topology-agnostic sampling reduces spatiotemporal redundancy but disrupts key topological connectivity, while the subsequent shallow and monolithic feature fusion inevitably leads to cross-modal contention. To overcome these limitations, we decouple the unified spatial reasoning task into two synergistic sub-problems. First, we formalize keyframe extraction as an information maximization problem on the spatiotemporal manifold under a sparsity constraint $N_n \ll N_k$. From the optimal sparse subset $\mathcal{V}_s \subset \mathcal{V}$, we extract their 2D visual features $\mathbf{F}_{\text{2D}}$, 3D geometric features $\mathbf{F}_{\text{3D}}$, and corresponding camera poses $\mathcal{P}$. Subsequently, we formalize multimodal alignment as a conditional routing policy. We introduce a routing function $\Phi_{\text{MoE}}$ that adaptively fuses these heterogeneous features, guided by the language instruction $\mathcal{Q}$ and spatial context from camera poses $\mathcal{P}$. The unified objective is to maximize the conditional likelihood of the target sequence:
\begin{equation}
    \mathcal{A}^* = \mathop{\arg\max}_{\mathcal{A}} \prod_{t=1}^T p_{\theta} \Big( a_t \mid a_{<t}, \Phi_{\text{MoE}} \big( \mathbf{F}_{\text{2D}}, \mathbf{F}_{\text{3D}} \mid \mathcal{P}, \mathcal{Q} \big), \mathcal{Q} \Big),
\end{equation}
where $p_{\theta}$ denotes the probability distribution parameterized by the generative language decoder $\mathcal{F}_{\text{LLM}}$ with learnable parameters $\theta$.

\begin{figure}[t]
    \centering
    \includegraphics[width=\textwidth]{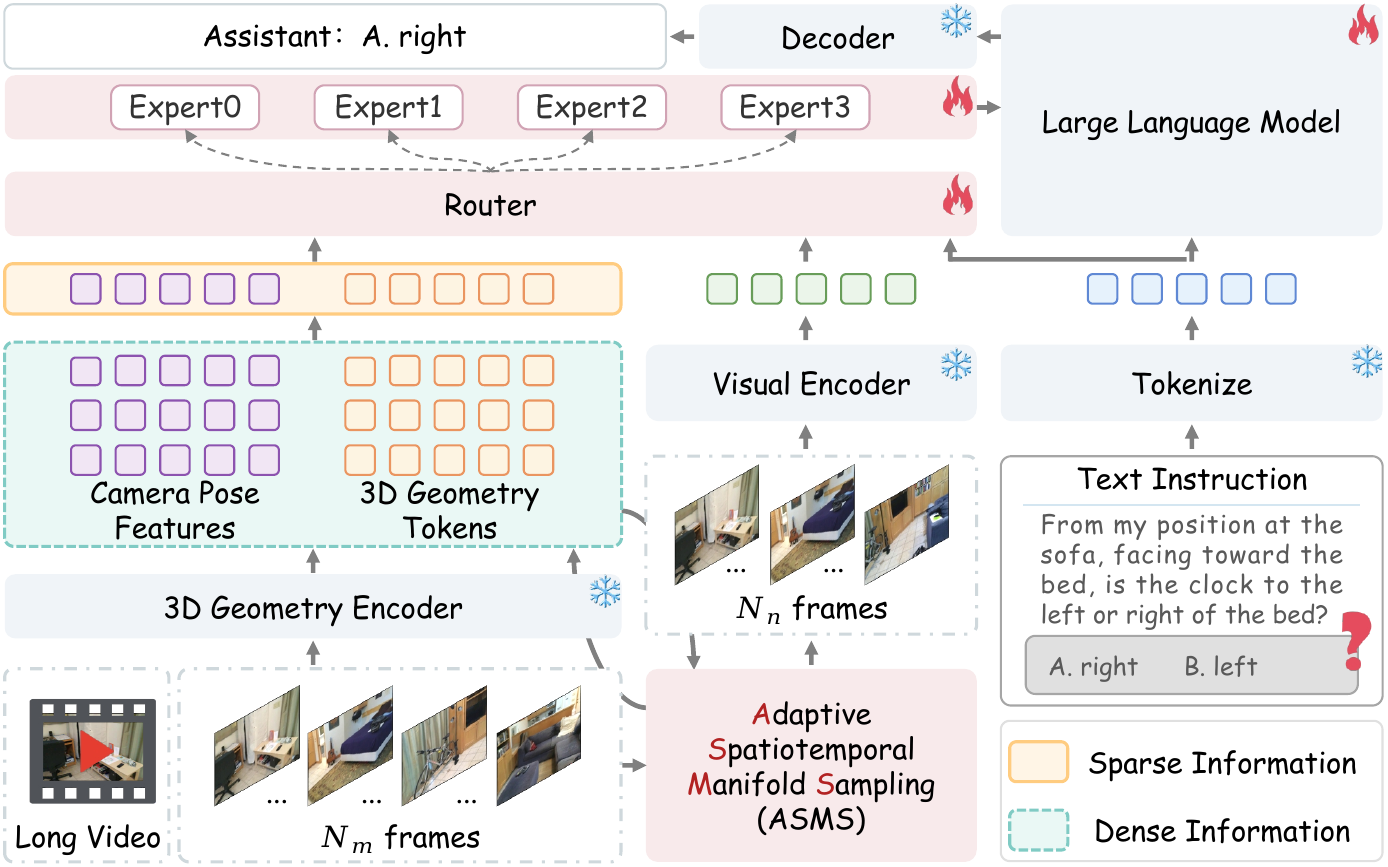}
    \caption{\textbf{Overview of SpaR3D-MoE framework.} 
    Given long videos, the ASMS constructs a spatiotemporal graph to adaptively distill $N_m$ candidates into $N_n$ informative keyframes. Along with text instructions, multimodal features are dynamically routed via the instruction-pose aware router to specialized experts ($E_0-E_3$), then aggregated to drive the MLLM for spatial reasoning.}
    \label{fig:architecture}
\end{figure}

Our end-to-end SpaR3D-MoE framework is shown in Fig.~\ref{fig:architecture}. First, long videos are sampled into sparse keyframes utilizing ASMS (Sec.~\ref{sec:sampling}). Next, visual and instruction tokens are encoded by Qwen3-VL~\cite{Bai2025Qwen3VLTR}, while 3D geometric and pose features are extracted by VGGT~\cite{wang2025vggt}. These features are then dynamically dispatched to four geometry-inductive experts via IPAR (Sec.~\ref{sec:MoE}). Finally, the adaptive multimodal features are projected into the LLM for autoregressive spatial reasoning, jointly optimized with a routing load-balancing penalty (Sec.~\ref{sec:opt_obj}).

\subsection{Adaptive Spatiotemporal Manifold Sampling}
\label{sec:sampling}
Continuous embodied observations naturally reside on a low-dimensional data manifold embedded within a high-dimensional multimodal feature space. 
To process these continuous streams with bounded computational overhead, we first uniformly downsample the $N_k$-frame video into $N_m=128$ candidates, preserving macroscopic scene topology. Since further rigid uniform sampling to achieve extreme sparsity ($N_n \ll N_m$, e.g., $N_n \in {8, 16, 32}$) disrupts this intrinsic connectivity, we introduce the ASMS mechanism, which extracts $N_n$ topologically crucial keyframes by approximating the manifold via a spatiotemporal graph.

To begin with, we establish a composite distance metric $\mathcal{D}(i, j)$ to quantify the frame-to-frame relations as the edge weights of our spatiotemporal graph. This metric integrates the relative camera displacement, implicit geometric dissimilarity, and temporal distance across any two frames:
\begin{equation}
\mathcal{D}(i, j) = \bar{\mathcal{D}}_{trans}(i, j) + \gamma \bar{\mathcal{D}}_{geo}(i, j) + \omega \bar{\mathcal{D}}_{tmp}(i, j),
\end{equation}
where $\bar{\mathcal{D}}_{trans}$ is the normalized $L_2$ distance of 3D translations from 6D poses predicted by VGGT~\cite{wang2025vggt}, serving as a spatial anchor to ensure broad trajectory coverage. To account for changes in viewing direction, $\bar{\mathcal{D}}_{geo}$ captures rotation-induced variations using cosine similarity of implicit 3D features encoded by VGGT, preventing redundant sampling at the same position. Meanwhile, $\bar{\mathcal{D}}_{tmp}$ uses normalized temporal interval to preserve temporal progression. We set $\gamma=0.5$ and $\omega=0.6$ to balance geometric diversity and temporal stability.

Building upon the distance metric, we introduce a node-level quality score $\mathcal{S}_i$ to ensure the informational validity of the sampled manifold. Relying solely on spatial distances risks anchoring the graph on uninformative regions, such as textureless blank walls. To address this, we quantify the visual richness of each candidate frame by computing the average $L_2$ norm of its Top-$K_v$ patch tokens $\mathbf{f}_{i,k}$. This prevents sparse foreground features from being diluted by massive background tokens. Additionally, to suppress motion blur caused by fast camera movement, we penalize this score with an exponential decay factor based on the instantaneous camera velocity $v_i$. The final quality score is defined as:
\begin{equation}
\mathcal{S}_i = \Bigg( \frac{1}{K_v} \sum_{k \in \mathcal{T}_{K_v}(i)} \|\mathbf{f}_{i,k}\|_2 \Bigg) \cdot \exp\left(-\frac{v_i}{\bar{v}}\right) + \epsilon ,
\end{equation}
where $\mathcal{T}_{K_v}(i)$ is the index set of the Top-$K_v$ tokens for frame $i$, $\bar{v}$ is the mean trajectory velocity, and $\epsilon$ ensures a baseline sampling probability.

Ultimately, to sparsify the dense video while preserving its underlying manifold structure, we formulate keyframe extraction as quality-gated farthest point sampling (FPS) process on the approximated spatiotemporal graph. At each iteration, we greedily select the candidate frame $i^*$ that maximizes its shortest distance to the already sampled subset $\mathcal{K}$, modulated by its quality score:
\begin{equation}
i^* = \arg\max_{i \notin \mathcal{K}} \Big[ \big( \min_{j \in \mathcal{K}} \mathcal{D}(i, j) \big) \cdot (1 + \lambda \mathcal{S}_i) \cdot \mathcal{M}(i) \Big],
\end{equation}
where $\lambda=3.0$ balances structural coverage and visual richness. To ensure topological robustness, we formulate $\mathcal{M}(i)$ as a motion-aware quality gate. Nodes scoring below $\tau = 0.6 \bar{\mathcal{S}}$ are penalized via $\mathcal{M}(i) = 0.1$, while reliable nodes retain $\mathcal{M}(i) = 1.0$. This effectively filters low-quality frames, yielding a sparse yet informative representation that covers the spatiotemporal manifold.

\begin{figure}[t]
    \centering
    \includegraphics[width=\textwidth]{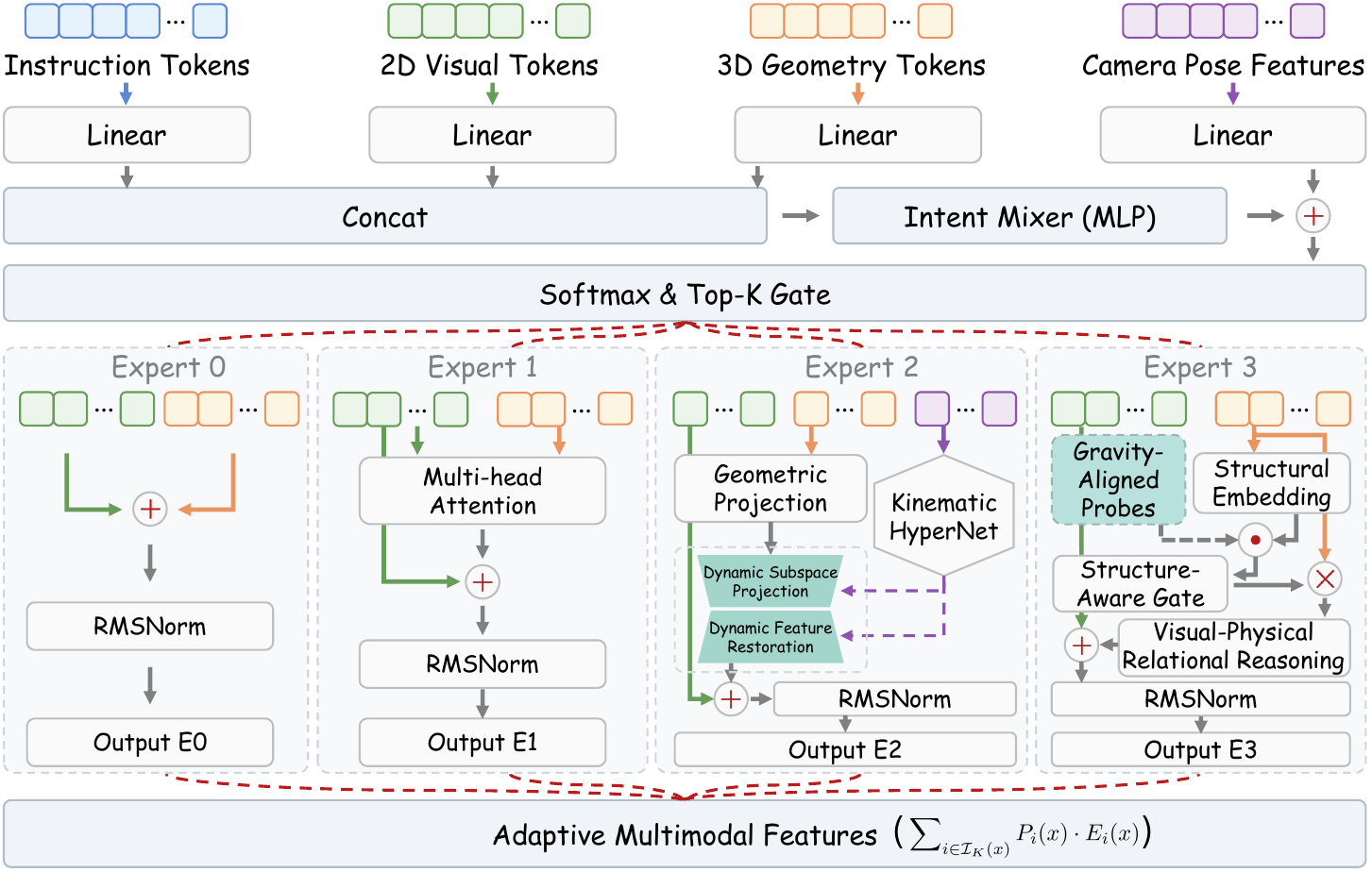}
    \caption{\textbf{Detailed architecture of the HGI-MoE.} The router processes multimodal features to generate routing probabilities, dynamically dispatching tokens to the most suitable experts ($E_0-E_3$), whose outputs are aggregated via a probability-weighted sum into adaptive multimodal features.}
    
    \label{fig:moe}
\end{figure}

\subsection{Heterogeneous Geometry-Inductive Mixture-of-Experts}
\label{sec:MoE}
\textbf{Instruction-Pose Aware Router.} As illustrated in Fig.~\ref{fig:moe}, our router jointly leverages the language instruction tokens $q$, 2D visual tokens $v$, 3D geometric tokens $g$, and camera pose features $p$. Specifically, the first three components ($q, v$, and $g$) are linearly projected and concatenated to capture task-specific multimodal correlations via an Intent Mixer (MLP). Meanwhile, to handle viewpoint variations caused by camera motion, the camera pose feature $p$ is linearly projected and added directly into the routing logit space as a global spatial bias. The routing logits $L \in \mathbb{R}^{N \times E}$ are computed as:
\begin{equation}
L = \text{MLP}([ \mathbf{W}_q q \,;\, \mathbf{W}_v v \,;\, \mathbf{W}_g g ]) + \mathbf{W}_p p,
\end{equation}
where $[ \cdot ; \cdot ]$ denotes channel-wise concatenation. This formulation ensures that the Top-$K$ expert activation is dynamically steered by both the user instruction and the underlying 3D spatial context.

\noindent\textbf{Heterogeneous Geometry-Inductive Experts.} As illustrated in the lower section of Fig.~\ref{fig:moe}, the routed multimodal tokens are dispatched to a specialized pool of four architecturally diversified experts. Each expert employs a tailored feature fusion mechanism to tackle distinct requirements of 3D spatial reasoning.

\textbf{Holistic Representation Expert ($E_0$).} While static shallow fusion often induces modal conflicts in monolithic architectures, we isolate this fundamental operation as a conditionally activated expert. $E_0$ achieves explicit modality integration via $E_0(v, g) = \text{RMSNorm}(v + g)$. Retaining this primitive baseline serves a dual purpose. It provides a natural contrast to our other specialized spatial experts and explicitly highlights the adaptive superiority of our MoE design. Unlike static monolithic fusion, this simple additive branch is dynamically assigned to tokens that do not require complex spatial transformations.

\textbf{Geometric-Semantic Cross-Attention Expert ($E_1$).} To establish a precise mapping between 2D visual semantics and 3D geometry, this expert employs a multi-head cross-attention mechanism. By using the visual tokens $v$ as queries and the geometric tokens $g$ as keys and values, $E_1$ explicitly injects 3D spatial priors into the 2D visual representations. Incorporating a residual connection, the aggregated output is formulated as $E_1(v, g) = \text{RMSNorm}(v + \text{CrossAttn}(Q=v, K=g, V=g))$. This adaptive alignment tightly couples 2D appearance with 3D geometry information, equipping the model with fine-grained spatial awareness for complex scene reasoning.

\textbf{Pose-Conditioned Dynamic Adapter Expert($E_2$).} To bridge the gap between static 3D geometry and large-baseline camera ego-motion across sparse views, $E_2$ operates as a pose-driven dynamic adapter. Rather than treating camera parameters as simple concatenated tokens, we leverage the view-specific pose features $p$ to actively modulate geometric features $g$ with view-dependent context. Using a HyperNet $\mathcal{H}$, the egocentric motion state is encoded into dynamic adaptation weights. Specifically, following an initial projection to obtain the base geometric embedding $g_{emb}$, these weights parametrize a low-rank bottleneck consisting of dynamic subspace projection and a subsequent feature restoration. The resulting pose-modulated output is then integrated into visual tokens $v$ via a learnable $\alpha$-scaled residual connection:

\begin{equation}
\begin{aligned}
    g_{adapted} &= \mathcal{F}_{\text{adapt}}(g_{emb} \mid \mathcal{H}(p)), \\
    E_2(v, g, p) &= \text{RMSNorm}(v + \alpha \cdot \text{Dropout}(g_{adapted})),
\end{aligned}
\label{eq:kama_adapter}
\end{equation}
where $\mathcal{F}_{\text{adapt}}$ denotes the rank-constrained dynamic transformation. This conditionally modulated design effectively transforms viewpoint-invariant geometric priors into a motion-aware feature space tailored for sparse visual inputs.

\textbf{Gravity-Aligned Structural Expert ($E_3$).} To ground physical metrics within complex 3D environments, $E_3$ establishes canonical references by identifying orthogonal structural foundations. Specifically, we introduce $N_p$ learnable structural probes $\mathbf{\Phi} \in \mathbb{R}^{N_p \times D}$ to capture gravity-aligned physical priors corresponding to the dominant scene layouts (e.g., floor planes). The geometric features $g$ are first projected into a structural subspace $g_{struct}$. A structure-aware mask $M$ is then derived from the cross-affinity between $g_{struct}$ and $\mathbf{\Phi}$ to filter out unstructured spatial noise. These gated geometric features $\tilde{g}$ are subsequently integrated with the visual tokens $v$ through a relational mapping network $\mathcal{F}_{\text{rel}}$:
\begin{equation}
\begin{aligned}
    M &= \sigma((g_{struct} \mathbf{\Phi}^T) \mathbf{W}_{attn}), \quad \tilde{g} = g \odot M, \\
    E_3(v, g) &= \text{RMSNorm}(v + \mathcal{F}_{\text{rel}}([v; \tilde{g}])),
\end{aligned}
\end{equation}
where $\mathbf{W}_{attn}$ is a dimension-matching projection, and $\sigma$ denotes the sigmoid activation. This design enables the model to encode global structural layouts, anchoring spatial reasoning in a consistent, gravity-aware coordinate system.

\textbf{Adaptive Multimodal Aggregation.} Given the routing logits $f(x)$, a Top-$K$ gating mechanism computes the adaptive probability distribution across the expert pool for each token $x$. The final multimodal representation is aggregated as a probability-weighted sum of the activated experts' outputs:
\begin{equation}
P_i(x) = \frac{e^{f(x)_i}}{\sum_{j \in \mathcal{I}_K(x)} e^{f(x)_j}}, \quad \text{MoE}(x) = \sum_{i \in \mathcal{I}_K(x)} P_i(x) \cdot E_i(x),
\end{equation}
where $\mathcal{I}_K(x)$ denotes the index set of the $K$ actively selected experts for token $x$. This adaptive activation ensures dynamic execution paths precisely tailored to the specific semantic queries and spatial complexities of the scene.

\subsection{Optimization Objectives}
\label{sec:opt_obj}
We train our framework end-to-end with a composite objective:
\begin{equation}
    \mathcal{L} = \mathcal{L}_{gen} + \lambda_{moe} \mathcal{L}_{moe},
\end{equation}
where $\mathcal{L}_{gen}$ is the standard cross-entropy loss for language generation, and $\lambda_{moe}$ is a scaling factor for the auxiliary routing penalty. Due to the sparse activation mechanism of MoE, expert utilization can become imbalanced during training, potentially leading to expert collapse. To address this, we introduce a load-balancing loss $\mathcal{L}_{moe}$ to encourage uniform expert usage across the token sequence. The load-balancing objective is defined as:
\begin{equation}
\mathcal{L}_{moe} = N_e \sum_{i=1}^{N_e} \bar{P}_i \cdot \rho_i,
\end{equation}
where $N_e$ is the total number of experts, the mean routing probability $\bar{P}_i$ and activation frequency $\rho_i$ for expert $i$ across all $N$ batch tokens are formulated as:
\begin{equation}
\bar{P}_i = \frac{1}{N} \sum_{t=1}^{N} P_i(x_t), \quad \rho_i = \frac{1}{N} \sum_{t=1}^{N} \mathbb{I} \left( i \in \mathcal{I}_K(x_t) \right),
\end{equation}
where $P_i(x_t)$ is the routing probability of expert $i$ for token $x_t$, $\mathcal{I}_K(x_t)$ denotes the $K$ highest-scoring expert indices, and $\mathbb{I}(\cdot)$ denotes the indicator function.

\section{Experiments}
To comprehensively evaluate SpaR3D-MoE across diverse 3D spatial tasks, we benchmark the model on three datasets that encompass varying levels of spatial reasoning complexity: VSI-Bench~\cite{yang2025thinking} for general spatial reasoning, ScanQA~\cite{azuma2022scanqa} for fine-grained scene question answering, and SQA3D~\cite{ma2023sqa3dsituatedquestionanswering} for situated reasoning. Due to space constraints, implementation details are provided in Appendix.

\subsection{Comparison with State-of-the-Art Methods}

\begin{table}[htbp]
  \centering
  \caption{\textbf{Comparison with SOTA methods on VSI-Bench.} We evaluate the Qwen3VL models using the lmms-eval~\cite{zhang2024lmmseval}. Notably, our SpaR3D-MoE achieves SOTA performance using solely 32 non-uniformly sampled sparse frames. Best results in each category are highlighted in \textbf{bold}, and the second-best are \underline{underlined}.}
  \label{tab:vsi_bench}
  
  \resizebox{\textwidth}{!}{
    \begin{tabular}{l | c | cccc | cccc}
      \toprule
      \multirow{2}{*}{\textbf{Methods}} & \multirow{2}{*}{\textbf{Avg.}} & \multicolumn{4}{c|}{\textbf{Numerical Answer Task}} & \multicolumn{4}{c}{\textbf{Multiple-Choice Answer Task}} \\
      \cmidrule(lr){3-6} \cmidrule(lr){7-10}
      & & Obj. Cnt. & Abs. Dist. & Obj. Size & Room Size & Rel. Dist. & Rel. Dir. & Route Plan & Appr. Order \\
      \midrule

      \rowcolor{tablegray}
      \multicolumn{10}{l}{\textit{Proprietary Models (API)}} \\
      GPT-4o~\cite{openai2024gpt4ocard} & 34.0 & 46.2 & 5.3 & 43.8 & 38.2 & 37.0 & 41.3 & 31.5 & 28.5 \\
      Gemini-1.5-Flash~\cite{geminiteam2024gemini15unlockingmultimodal} & 42.1 & 49.8 & 30.8 & 53.5 & 54.4 & 37.7 & 41.0 & 31.5 & 37.8 \\
      Gemini-1.5-Pro~\cite{geminiteam2024gemini15unlockingmultimodal} & 45.4 & 56.2 & 30.9 & 64.1 & 43.6 & 51.3 & 46.3 & 36.0 & 34.6 \\

      \addlinespace[0.3em]

      \rowcolor{tablegray}
      \multicolumn{10}{l}{\textit{Open-source Models}} \\
      InternVL2-8B~\cite{chen2024internvl} & 34.6 & 23.1 & 28.7 & 48.2 & 39.8 & 36.7 & 30.7 & 29.9 & 39.6 \\
      InternVL2-40B~\cite{chen2024internvl} & 36.0 & 34.9 & 26.9 & 46.5 & 31.8 & 42.1 & 32.2 & 34.0 & 39.6 \\
      InternVL3-78B~\cite{zhu2025internvl3exploringadvancedtraining} & 48.5 & \textbf{71.2} & \textbf{53.7} & 44.4 & 39.5 & \textbf{55.9} & 39.5 & 28.9 & 54.5 \\
      LongVILA-8B~\cite{chen2024longvilascalinglongcontextvisual} & 21.6 & 29.1 & 9.1 & 16.7 & 0.0 & 29.6 & 30.7 & 32.5 & 25.5 \\
      VILA-1.5-40B~\cite{lin2024vila} & 31.2 & 22.4 & 24.8 & 48.7 & 22.7 & 40.5 & 25.7 & 31.5 & 32.9 \\
      LongVA-7B~\cite{Zhang2024LongCT} & 29.2 & 38.0 & 16.6 & 38.9 & 22.2 & 33.1 & 43.3 & 25.4 & 15.7 \\
      LLaVA-NeXT-Video-7B~\cite{zhang2024llavanextvideo} & 35.6 & 48.5 & 14.0 & 47.8 & 24.2 & 43.5 & 42.4 & 34.0 & 30.6 \\
      LLaVA-NeXT-Video-72B~\cite{zhang2024llavanextvideo} & 40.9 & 48.9 & 22.8 & 57.4 & 35.3 & 42.4 & 36.7 & \textbf{35.0} & 48.6 \\
      LLaVA-OneVision-7B~\cite{li2024llavaonevisioneasyvisualtask} & 32.4 & 47.7 & 20.2 & 47.4 & 12.3 & 42.5 & 35.2 & 29.4 & 24.4 \\
      LLaVA-OneVision-72B~\cite{li2024llavaonevisioneasyvisualtask} & 40.2 & 43.5 & 23.9 & 57.6 & 37.5 & 42.5 & 39.9 & 32.5 & 44.6 \\
      Qwen2.5VL-7B~\cite{Bai2025Qwen25VLTR} & 33.0 & 40.9 & 14.8 & 43.4 & 10.7 & 38.6 & 38.5 & 33.0 & 29.8 \\
      Qwen3VL-4B~\cite{Bai2025Qwen3VLTR} & \underline{54.8} & 66.4 & \underline{43.4} & \textbf{74.2} & \underline{60.0} & 53.0 & \underline{46.2} & 32.0 & \underline{63.1} \\
      Qwen3VL-8B~\cite{Bai2025Qwen3VLTR} & \textbf{55.7} & \underline{68.0} & 45.8 & \underline{73.5} & \textbf{60.3} & \underline{53.7} & \textbf{46.3} & \underline{32.5} & \textbf{65.5} \\

      \addlinespace[0.3em]

      \rowcolor{tablegray}
      \multicolumn{10}{l}{\textit{Spatial-Aware MLLMs}} \\
      VG LLM-4B~\cite{Zheng2025LearningFV} & 46.1 & \underline{66.4} & \underline{36.6} & 55.2 & \underline{56.3} & 40.8 & 43.4 & 30.4 & 39.5 \\
      Spacer~\cite{ouyang2025spacerreinforcingmllmsvideo} & 45.5 & 57.8 & 28.2 & 59.9 & 47.1 & 40.1 & 45.4 & \underline{33.5} & \underline{52.1} \\
      ViLaSR~\cite{wu2025reinforcingspatialreasoningvisionlanguage} & 45.4 & 63.5 & 34.4 & 60.6 & 30.9 & \underline{48.9} & 45.2 & 30.4 & 49.2 \\
      Spatial-MLLM~\cite{Wu2025SpatialMLLMBM} & \underline{48.4} & 65.3 & 34.8 & \underline{63.1} & 45.1 & 41.3 & \underline{46.2} & \underline{33.5} & 46.3 \\
      
      \rowcolor{myblue} \textbf{SpaR3D-MoE (Ours)} & \textbf{63.5} & \textbf{71.3} & \textbf{48.0} & \textbf{72.8} & \textbf{69.6} & \textbf{58.7} & \textbf{70.1} & \textbf{44.0} & \textbf{73.5} \\
      \bottomrule
    \end{tabular}
  }
\end{table}

\textbf{Evaluation on VSI-Bench.} 
As summarized in Table~\ref{tab:vsi_bench}, SpaR3D-MoE achieves a new SOTA with a 63.5 average on VSI-Bench, surpassing the strongest baseline Qwen3VL-8B by 7.8 points, while yielding relative gains of 35.4\% in Route Plan and 51.4\% in Relative Direction. Notably, our framework also exceeds the leading commercial model, Gemini-1.5-Pro, by a margin of 18.1 points. This performance gap is particularly evident in complex tasks, such as Route Plan (44.0 vs. 36.0) and Relative Direction (70.1 vs. 46.3). Moreover, SpaR3D-MoE achieves these advantages utilizing only 32 non-uniformly sampled sparse frames, in contrast to the dense input required by Gemini-1.5-Pro ($\sim$85 frames). We attribute these gains to our ASMS mechanism, which adaptively selects high-quality sparse keyframes from long video sequences to construct a comprehensive and informative scene context. Building upon this, our HGI-MoE architecture employs a dynamic router guided by instructions and camera poses to dispatch input tokens to dedicated experts. This mechanism ensures an adaptive and comprehensive fusion of multimodal features, driving performance improvements across diverse complex spatial reasoning tasks.

\begin{table}[t]
\centering
\caption{\textbf{Evaluations on the ScanQA (val)}. B-1 to B-4 are BLEU-n scores. Best results in each category are highlighted in \textbf{bold}, and the second-best are \underline{underlined}.}
\label{tab:scanqa}
\resizebox{\textwidth}{!}{
\setlength{\tabcolsep}{7pt}
\begin{tabular}{l | c c c c c c c c}
\toprule
\multirow{2}{*}[-2pt]{\textbf{Methods}} & \multicolumn{8}{c}{\textbf{ScanQA (val)}} \\
\cmidrule{2-9}
& EM@1 & B-1 & B-2 & B-3 & B-4 & ROUGE-L & METEOR & CIDEr \\
\midrule
\rowcolor{tablegray} \multicolumn{9}{l}{\textit{Task-Specific Models}} \\
ScanQA~\cite{azuma2022scanqa} & 21.1 & 30.2 & 20.4 & 15.1 & 10.1 & 33.3 & 13.1 & 64.9 \\
3D-Vista~\cite{zhu20233dvista} & 22.4 & - & - & - & 10.4 & 35.7 & 13.9 & 69.6 \\

\addlinespace[0.4em]

\rowcolor{tablegray} \multicolumn{9}{l}{\textit{3D/2.5D-Input Models}} \\
3D-LLM~\cite{chen2024grounded-3d-llm} & 20.5 & 39.3 & 25.2 & 18.4 & 12.0 & 35.7 & 14.5 & 69.4 \\
LL3DA~\cite{chen2024ll3da} & - & - & - & - & 13.5 & 37.3 & 15.9 & 76.8 \\
Chat-Scene~\cite{huang2024chatscene} & 21.6 & 43.2 & 29.1 & 20.6 & 14.3 & 41.6 & 18.0 & 87.7 \\
3D-LLaVA~\cite{deng20253d} & - & - & - & - & \underline{17.1} & \underline{43.1} & \underline{18.4} & \underline{92.6} \\
Video-3D LLM~\cite{zheng2025video} & \textbf{30.1} & \textbf{47.1} & \textbf{31.7} & \textbf{22.8} & \textbf{16.2} & \textbf{49.0} & \textbf{19.8} & \textbf{102.1} \\

\addlinespace[0.4em]

\rowcolor{tablegray} \multicolumn{9}{l}{\textit{Video-Input Models}} \\
Qwen2.5-VL-3B~\cite{Bai2025Qwen25VLTR} & 15.4 & 22.5 & 13.1 & 8.1 & 3.8 & 25.4 & 9.7 & 47.4 \\
Qwen2.5-VL-7B~\cite{Bai2025Qwen25VLTR} & 19.0 & 27.8 & 13.6 & 6.3 & 3.0 & 29.3 & 11.4 & 53.9 \\
Qwen2.5-VL-72B~\cite{Bai2025Qwen25VLTR} & 24.0 & 26.8 & 17.8 & 14.6 & 12.0 & 35.2 & 13.0 & 66.9 \\
LLaVA-Video-7B~\cite{zhang2024llavanextvideo} & - & 39.7 & 26.6 & 9.3 & 3.1 & 44.6 & 17.7 & 88.7 \\
Oryx-34B~\cite{liu2025oryxmllmondemandspatialtemporal} & - & 38.0 & 24.6 & - & - & 37.3 & 15.0 & 72.3 \\
Spatial-MLLM~\cite{Wu2025SpatialMLLMBM} & \underline{26.3} & \underline{44.4} & \underline{28.8} & \underline{21.9} & \underline{14.8} & \underline{45.0} & \underline{18.4} & \underline{91.8} \\

\rowcolor{myblue} \textbf{SpaR3D-MoE (Ours)} & \textbf{30.4} & \textbf{46.4} & \textbf{31.6} & \textbf{23.3} & \textbf{17.1} & \textbf{48.6} & \textbf{19.5} & \textbf{101.5} \\
\bottomrule
\end{tabular}
}
\end{table}

\textbf{Evaluation on ScanQA.} 
As detailed in Tab.~\ref{tab:scanqa}, SpaR3D-MoE performs competitively on the ScanQA validation set, a benchmark that requires both semantic grounding and spatial reasoning. Our framework achieves SOTA performance among video-based models using sparse RGB inputs, reaching an EM@1 of 30.4 and a CIDEr of 101.5. Specifically, it surpasses Video-3D LLM in Exact Match (30.4 vs. 30.1) while exceeding 3D-LLaVA by a significant margin in CIDEr (101.5 vs. 92.6). These results demonstrate that by dispatching multimodal features to suitable experts guided by instructions and camera poses, our approach seamlessly adapts to diverse task instructions, achieving robust spatial understanding from sparse RGB keyframes without costly 3D-specific data.

\begin{table}[t]
\centering
\caption{\textbf{Evaluations on the SQA3D (test).} All scores are reported in EM@1. Best results in each category are highlighted in \textbf{bold}, and the second-best are \underline{underlined}.}
\label{tab:sqa3d}
\resizebox{\textwidth}{!}{
\setlength{\tabcolsep}{8pt}
\begin{tabular}{l | c c c c c c c}
\toprule
\multirow{2}{*}[-2pt]{\textbf{Methods}} & \multicolumn{7}{c}{\textbf{SQA3D (test)}} \\
\cmidrule{2-8}
& What & Is & How & Can & Which & Others & Avg. \\
\midrule
\rowcolor{tablegray} \multicolumn{8}{l}{\textit{Task-Specific Models}} \\
SQA3D~\cite{ma2023sqa3dsituatedquestionanswering} & 31.6 & 63.8 & 46.0 & 69.5 & 43.9 & 45.3 & 46.6 \\
3D-Vista~\cite{zhu20233dvista} & 34.8 & 63.3 & 45.4 & 69.8 & 47.2 & 48.1 & 48.5 \\

\addlinespace[0.4em]

\rowcolor{tablegray} \multicolumn{8}{l}{\textit{3D/2.5D-Input Models}} \\
Scene-LLM~\cite{fu2024scene-llm} & 40.9 & \underline{69.1} & 45.0 & \textbf{70.8} & 47.2 & 52.3 & 54.2 \\
Chat-Scene~\cite{huang2024chatscene} & \underline{45.4} & 67.0 & \underline{52.0} & 69.5 & \underline{49.9} & \underline{55.0} & \underline{54.6} \\
Video-3D LLM~\cite{zheng2025video} & \textbf{51.1} & \textbf{72.4} & \textbf{55.5} & \underline{69.8} & \textbf{51.3} & \textbf{56.0} & \textbf{58.6} \\

\addlinespace[0.4em]

\rowcolor{tablegray} \multicolumn{8}{l}{\textit{Video-Input Models}} \\
Qwen2.5-VL-3B~\cite{Bai2025Qwen25VLTR} & 34.8 & 52.1 & 39.8 & 52.7 & 45.6 & 47.0 & 43.4 \\
Qwen2.5-VL-7B~\cite{Bai2025Qwen25VLTR} & 39.7 & 56.6 & 41.1 & 55.9 & 47.6 & 47.2 & 46.5 \\
Qwen2.5-VL-72B~\cite{Bai2025Qwen25VLTR} & 41.7 & 56.3 & 41.5 & 55.6 & 44.5 & 48.0 & 47.0 \\
LLaVA-Video-7B~\cite{zhang2024llavanextvideo} & 42.7 & 56.3 & 47.5 & 55.3 & 50.1 & 47.2 & 48.5 \\
Spatial-MLLM-4B~\cite{Wu2025SpatialMLLMBM} & \underline{45.9} & \underline{71.6} & \underline{55.1} & \textbf{69.5} & \underline{52.0} & \underline{53.0} & \underline{55.9} \\

\rowcolor{myblue} \textbf{SpaR3D-MoE (Ours)} & \textbf{51.6} & \textbf{72.6} & \textbf{56.3} & \underline{69.2} & \textbf{52.1} & \textbf{54.2} & \textbf{58.3} \\
\bottomrule
\end{tabular}
}
\end{table}

\textbf{Evaluation on SQA3D.}
Beyond static scene understanding, we evaluate SpaR3D-MoE on the SQA3D benchmark to assess its situated reasoning capabilities. As reported in Tab.~\ref{tab:sqa3d}, our framework establishes a new SOTA among video-based methods with an average EM@1 of 58.3. Notably, it even outperforms the explicit 3D-based Video-3D LLM in fine-grained categories like \textit{What} and \textit{How}. We attribute this competitive performance to a synergistic design where our ASMS provides rich scene context while preserving topological connectivity, establishing an informative foundation. Furthermore, the HGI-MoE adaptively activates specialized experts guided by situated instructions and camera poses, effectively leveraging multimodal features to achieve precise and robust spatial understanding from RGB-only inputs.

\subsection{Ablation Study}
To validate the architectural designs of SpaR3D-MoE, we perform comprehensive ablation studies on VSI-Bench~\cite{yang2025thinking}. We systematically analyze the contributions of heterogeneous experts, the role of multimodal routing guidance, and the impact of our adaptive sampling mechanism across different frame densities.

\begin{table}[t]
\centering
\caption{\textbf{Ablation on expert roles and routing guidance.} We analyze each expert ($E_0-E_3$) via selective masking and evaluate multi-modal routing inputs, where the Base Router relies exclusively on visual and geometric features.}
\label{tab:ablation_merged}
\resizebox{\textwidth}{!}{
\begin{tabular}{l|c|cccc|cccc}
\toprule
\multirow{2}{*}{\textbf{Model Variant}} & \multirow{2}{*}{\textbf{Avg.}} & \multicolumn{4}{c|}{\textbf{Numerical Answer Tasks}} & \multicolumn{4}{c}{\textbf{Multiple-Choice Answer Tasks}} \\
\cmidrule(lr){3-6} \cmidrule(l){7-10}
 & & Obj. Cnt. & Abs. Dist. & Obj. Size & Room Size & Rel. Dist. & Rel. Dir. & Route Plan & Appr. Order \\
\midrule
\rowcolor{gray!15} \multicolumn{10}{l}{\textit{Ablation on Expert Roles}} \\
$w/o$ $E_0$ & 61.8 & 71.2 & 46.3 & 72.9 & 68.7 & 57.5 & 65.6 & 41.3 & 70.9 \\
$w/o$ $E_1$ & 62.8 & 70.5 & 47.3 & 72.0 & 69.1 & 57.3 & 69.7 & 43.0 & 73.5 \\
$w/o$ $E_2$ & 59.3 & 69.8 & 42.2 & 72.2 & 63.0 & 55.7 & 60.4 & 38.8 & 72.3 \\
$w/o$ $E_3$ & 62.4 & 69.8 & 47.5 & 71.7 & 68.8 & 58.0 & 68.6 & 40.0 & 74.8 \\
\midrule
\rowcolor{gray!15} \multicolumn{10}{l}{\textit{Ablation on Routing Guidance}} \\
Base Router & 61.2 & 70.5 & 46.0 & 71.5 & 67.5 & 56.5 & 67.5 & 39.0 & 71.1 \\
+ Pose Bias ($p$) & 62.5 & 70.8 & 47.2 & 72.2 & 68.8 & 57.8 & 68.5 & 42.0 & 72.7 \\
+ Query Guidance ($q$) & 61.8 & 70.3 & 46.5 & 71.9 & 68.0 & 57.2 & 66.0 & 41.5 & 73.0 \\
\midrule
\rowcolor{myblue} \textbf{SpaR3D-MoE (Ours)} & 63.5 & 71.3 & 48.0 & 72.8 & 69.6 & 58.7 & 70.1 & 44.0 & 73.5 \\
\bottomrule
\end{tabular}
}
\end{table}

\textbf{Effectiveness of Heterogeneous Geometry-Inductive Experts.} 
To evaluate expert specialization within HGI-MoE, we selectively mask individual experts during inference to analyze their performance impacts, as detailed in Tab.~\ref{tab:ablation_merged}.
First, masking the \textbf{holistic representation expert ($E_0$)} reduces overall performance by 1.7. Instead of uniformly affecting all tasks, it significantly impairs Relative Direction and Appearance Order by 4.5 and 2.6, respectively. This validates our motivation to isolate simple additive fusion. Without $E_0$, tokens not requiring complex spatial processing are routed through high-order experts, resulting in over-processing of raw features and diminishing the specialized capacity of other branches. 
Furthermore, masking the \textbf{geometric-semantic cross-attention expert ($E_1$)} specifically degrades fine-grained spatial relational tasks, dropping Relative Distance and Object Counting by 1.4 and 0.8. This underscores the essential role of explicit cross-attention between 2D visual queries and 3D geometric keys for precise multi-object spatial grounding in 3D scene understanding. 
Crucially, masking the \textbf{pose-conditioned dynamic adapter expert ($E_2$)} causes the most severe overall degradation (a 4.2 drop), with Relative Direction and Route Plan declining by 9.7 and 5.2, respectively. This confirms $E_2$'s essential function in leveraging camera pose to mitigate spatial misalignment from sparse viewpoints, thereby acting as an implicit coordinate transformer to align these disjointed observations. 
Finally, masking the \textbf{gravity-aligned structural expert ($E_3$)} severely impairs spatial topology and navigation tasks, with Route Plan dropping by 4.0. This demonstrates that our learnable gravity-aligned probes effectively capture the structural foundations and physical anchors necessary for robust spatial grounding. Collectively, these ablation results confirm that each expert fulfills a distinct, specialized role, jointly enhancing the model's comprehensive 3D spatial understanding.

\noindent \textbf{Significance of Multimodal Routing Guidance.} 
Building upon foundational visual and geometric features, our IPAR introduces task instructions and camera poses to drive cross-modal interactions for precise expert selection. As shown in the lower section of Tab.~\ref{tab:ablation_merged}, integrating these additional modalities significantly optimizes routing decisions. Compared to the base router, task instructions provide semantic guidance with a 0.6 average improvement, while incorporating camera poses yields a larger gain of 1.3. Specifically, pose features directly benefit viewpoint-dependent tasks, increasing Route Plan and Relative Direction by 3.0 and 1.0, respectively. This underscores both modalities as essential condition priors. Their joint synergy culminates in the peak average of 63.5, demonstrating that precise expert dispatching relies on tightly coupling task-aware semantic intent with pose dynamics.

\begin{table}[t]
\centering
\caption{\textbf{Ablation on sampling strategy and frame density.} We compare ASMS against uniform sampling across varying frame counts on VSI-Bench. ASMS consistently achieves higher performance, with 32 frames delivering the best results.}
\label{tab:ablation_frame_count}
\resizebox{\textwidth}{!}{
\begin{tabular}{l|c|cccc|cccc}
\toprule
\multirow{2}{*}{\textbf{\shortstack{Sampling \\ Strategy}}} & \multirow{2}{*}{\textbf{Avg.}} & \multicolumn{4}{c|}{\textbf{Numerical Answer Tasks}} & \multicolumn{4}{c}{\textbf{Multiple-Choice Answer Tasks}} \\
\cmidrule(lr){3-6} \cmidrule(l){7-10}
 & & Obj. Cnt. & Abs. Dist. & Obj. Size & Room Size & Rel. Dist. & Rel. Dir. & Route Plan & Appr. Order \\
\midrule
\rowcolor{gray!15} \multicolumn{10}{l}{\textit{Uniform Sampling}} \\
8 Frames  & 56.6 & 68.2 & 42.1 & 70.0 & 65.1 & 53.8 & 60.1 & 34.0 & 59.5 \\
16 Frames & 61.2 & 70.7 & 47.1 & 71.8 & 68.7 & 57.9 & 68.1 & 38.0 & 67.3 \\
32 Frames & 62.4 & 71.5 & 47.8 & 72.6 & 68.9 & 58.2 & 68.4 & 39.8 & 72.0 \\
\midrule
\rowcolor{gray!15} \multicolumn{10}{l}{\textit{ASMS (Ours)}} \\
8 Frames  & 57.8 & 68.5 & 42.6 & 70.2 & 65.5 & 54.3 & 62.3 & 37.6 & 61.5 \\
16 Frames & 61.9 & 70.8 & 47.5 & 71.5 & 69.3 & 58.3 & 69.2 & 40.1 & 68.5 \\
\rowcolor{myblue} 32 Frames & 63.5 & 71.3 & 48.0 & 72.8 & 69.6 & 58.7 & 70.1 & 44.0 & 73.5 \\
\bottomrule
\end{tabular}
}
\vspace{-4mm}
\end{table}

\textbf{Impact of Sampling Strategy and Frame Density.} 
To evaluate our keyframe extraction mechanism, Tab.~\ref{tab:ablation_frame_count} compares the ASMS mechanism against uniform sampling across varying frame counts. ASMS consistently outperforms the uniform baseline across all configurations, demonstrating its ability to capture critical spatial structures for 3D understanding. Notably, the 32-frame setup achieves the highest average score of 63.5, highlighted by a 10.6\% improvement in the complex Route Plan task. Even with only 16 frames, the model maintains a competitive performance of 61.9, confirming that ASMS effectively filters spatiotemporal redundancy. Although extreme sparsity at 8 frames leads to an overall performance drop, ASMS exhibits greater robustness on complex tasks like Route Plan and Appearance Order. These results indicate that our adaptive sampling successfully retains spatially informative frames, ensuring reliable 3D reasoning even under highly sparse conditions.

\section{Conclusion}
We introduced SpaR3D-MoE, a novel framework that endows MLLMs with physically-grounded spatial intelligence relying solely on sparse RGB views, obviating the need for 3D-specific data. By sampling sparse keyframes using the proposed ASMS mechanism, the model effectively filters spatiotemporal redundancy while preserving essential scene topology. Building upon this foundation, the introduced HGI-MoE adaptively dispatches multimodal tokens to specialized experts with distinct, tailored cross-modal fusion capacities, fulfilling the diverse requirements of 3D spatial reasoning. Extensive experiments demonstrate that SpaR3D-MoE achieves SOTA performance on the challenging VSI-Bench, ScanQA, and SQA3D benchmarks, confirming the effectiveness and generalizability of our approach across comprehensive spatial understanding tasks. Future work will extend this framework to online video stream spatial reasoning.

\section{Acknowledgments}
This work was supported by the project ``Research of Rapid 3D Digital Acquisition and Reconstruction Technology and Equipment for Cultural Heritage'' (No.2024JK4002) and the National Natural
Science Foundation of China under Grant Nos. 62572468 and
62402493.

%
%
\bibliographystyle{splncs04}
\bibliography{main}

@String(PAMI  = {IEEE Trans. Pattern Anal. Mach. Intell.})

@String(CVPR  = {IEEE Conf. Comput. Vis. Pattern Recog.})

@String(ICCV  = {Int. Conf. Comput. Vis.})

@String(NeurIPS = {Adv. Neural Inform. Process. Syst.})

@String(ICML  = {Int. Conf. Mach. Learn.})

@String(TMLR  = {Trans. Mach. Learn Res.})

@inproceedings{anderson2018vision,
  title={Vision-and-language navigation: Interpreting visually-grounded navigation instructions in real environments},
  author={Anderson, Peter and Wu, Qi and Teney, Damien and Bruce, Jake and Johnson, Mark and S{\"u}nderhauf, Niko and Reid, Ian and Gould, Stephen and Van Den Hengel, Anton},
  booktitle=CVPR,
  pages={3674--3683},
  year=2018
}

@inproceedings{zhi2025lscenellm,
  title={Lscenellm: Enhancing large 3d scene understanding using adaptive visual preferences},
  author={Zhi, Hongyan and Chen, Peihao and Li, Junyan and Ma, Shuailei and Sun, Xinyu and Xiang, Tianhang and Lei, Yinjie and Tan, Mingkui and Gan, Chuang},
  booktitle=CVPR,
  pages={3761--3771},
  year={2025}
}

@inproceedings{alayrac2022flamingo,
  title={Flamingo: a visual language model for few-shot learning},
  author={Alayrac, Jean-Baptiste and Donahue, Jeff and Luc, Pauline and Miech, Antoine and Barr, Iain and Hasson, Yana and Lenc, Karel and Mensch, Arthur and Millican, Katherine and Reynolds, Malcolm and others},
  booktitle=NeurIPS,
  volume={35},
  year={2022}
}

@inproceedings{liu2023visual,
  title={Visual instruction tuning},
  author={Liu, Haotian and Li, Chunyuan and Wu, Qingyang and Lee, Yong Jae},
  booktitle=NeurIPS,
  volume={36},
  pages={34892--34916},
  year={2023}
}

@article{Bai2025Qwen25VLTR,
  title={Qwen2.5-VL Technical Report},
  author={Shuai Bai and Keqin Chen and Xuejing Liu and Jialin Wang and Wenbin Ge and Sibo Song and Kai Dang and Peng Wang and Shijie Wang and Jun Tang and others},
  journal={ArXiv},
  year={2025},
  volume={abs/2502.13923}
}

@inproceedings{qi2017pointnet++,
  title={Pointnet++: Deep hierarchical feature learning on point sets in a metric space},
  author={Qi, Charles Ruizhongtai and Yi, Li and Su, Hao and Guibas, Leonidas J},
  booktitle=NeurIPS,
  volume={30},
  pages={5099--5108},
  year={2017}
}

@article{ungerleider1994and,
  title={‘What’and ‘where’in the human brain},
  author={Ungerleider, Leslie G and Haxby, James V},
  journal={Current opinion in neurobiology},
  volume={4},
  number={2},
  pages={157--165},
  year={1994},
  publisher={Elsevier}
}

@inproceedings{Tong2024Cambrian1AF,
  title={Cambrian-1: A Fully Open, Vision-Centric Exploration of Multimodal LLMs},
  author={Shengbang Tong and Ellis Brown and Penghao Wu and Sanghyun Woo and Manoj Middepogu and Sai Charitha Akula and Jihan Yang and Shusheng Yang and Adithya Iyer and Xichen Pan and Austin Wang and Rob Fergus and Yann LeCun and Saining Xie},
  booktitle=NeurIPS,
  volume={37},
  year={2024}
}

@inproceedings{chen2024spatialvlm,
  title={Spatialvlm: Endowing vision-language models with spatial reasoning capabilities},
  author={Chen, Boyuan and Xu, Zhuo and Kirmani, Sean and Ichter, Brain and Sadigh, Dorsa and Guibas, Leonidas and Xia, Fei},
  booktitle=CVPR,
  pages={14455--14465},
  year={2024}
}

@article{Tang2025LEGOPuzzlesHG,
  title={LEGO-Puzzles: How Good Are MLLMs at Multi-Step Spatial Reasoning?},
  author={Kexian Tang and Junyao Gao and Yanhong Zeng and Haodong Duan and Yanan Sun and Zhening Xing and Wenran Liu and Kaifeng Lyu and Kai Chen},
  journal={ArXiv},
  year={2025},
  volume={abs/2503.19990}
}

@article{Bai2025Qwen3VLTR,
  title={Qwen3-VL Technical Report},
  author={Shuai Bai and Yuxuan Cai and Ruizhe Chen and Keqin Chen and others},
  journal={ArXiv},
  year={2025},
  volume={abs/2511.21631}
}

@inproceedings{hong20233d,
  title={3D-LLM: Injecting the 3D World into Large Language Models},
  author={Hong, Yining and Zhen, Haoyu and Chen, Peihao and Zheng, Shuhong and Du, Yilun and Chen, Zhenfang and Gan, Chuang},
  booktitle=NeurIPS,
  volume={36},
  address={New Orleans, LA},
  pages={20482--20494},
  year={2023}
}

@article{wang2023chat-3d,
    title={Chat-3D: Data-efficiently Tuning Large Language Model for Universal Dialogue of 3D Scenes},
    author={Zehan Wang and Haifeng Huang and Yang Zhao and Ziang Zhang and Zhou Zhao},
    journal={ArXiv},
    year={2023},
    volume={abs/2308.08769}
}

@inproceedings{huang2024chatscene,
  title={Chat-scene: Bridging 3d scene and large language models with object identifiers},
  author={Huang, Haifeng and Chen, Yilun and Wang, Zehan and Huang, Rongjie and Xu, Runsen and Wang, Tai and Liu, Luping and Cheng, Xize and Zhao, Yang and Pang, Jiangmiao and others},
  booktitle=NeurIPS,
  volume={37},
  address = {Vancouver, BC, Canada},
  year={2024}
}

@article{fu2024scene-llm,
    title={Scene-LLM: Extending Language Model for 3D Visual Understanding and Reasoning},
    author={Rao Fu and Jingyu Liu and Xilun Chen and Yixin Nie and Wenhan Xiong},
    journal={ArXiv},
    year={2024},
    volume={abs/2403.11401}
}

@inproceedings{zhu2024llava-3d,
    title={LLaVA-3D: {A} Simple Yet Effective Pathway to Empowering LMMs with 3D Capabilities},
    author={Chenming Zhu and Tai Wang and Wenwei Zhang and Jiangmiao Pang and Xihui Liu},
    booktitle=ICCV,
    pages={4295--4305},
    year={2025},
    doi={10.1109/ICCV51701.2025.00409}
}

@article{chen2024grounded-3d-llm,
    title={Grounded 3D-LLM with Referent Tokens},
    author={Yilun Chen and Shuai Yang and Haifeng Huang and Tai Wang and Runsen Xu and Ruiyuan Lyu and Dahua Lin and Jiangmiao Pang},
    journal={ArXiv},
    year={2024},
    volume={abs/2405.10370}
}

@inproceedings{huang2024LEO,
  title={An Embodied Generalist Agent in 3D World},
  author={Huang, Jiangyong and Yong, Silong and Ma, Xiaojian and Linghu, Xiongkun and Li, Puhao and Wang, Yan and Li, Qing and Zhu, Song-Chun and Jia, Baoxiong and Huang, Siyuan},
  booktitle=ICML,
  pages={20413--20451},
  volume={235},
  year={2024}
}

@article{Zheng2025LearningFV,
  title={Learning from Videos for 3D World: Enhancing MLLMs with 3D Vision Geometry Priors},
  author={Duo Zheng and Shijia Huang and Yanyang Li and Liwei Wang},
  journal={ArXiv},
  year={2025},
  volume={abs/2505.24625}
}

@article{Wu2025SpatialMLLMBM,
  title={Spatial-MLLM: Boosting MLLM Capabilities in Visual-based Spatial Intelligence},
  author={Diankun Wu and Fangfu Liu and Yi-Hsin Hung and Yueqi Duan},
  journal={ArXiv},
  year={2025},
  volume={abs/2505.23747}
}

@article{ma2023sqa3dsituatedquestionanswering,
  title={SQA3D: Situated Question Answering in 3D Scenes},
  author={Xiaojian Ma and Silong Yong and Zilong Zheng and Qing Li and Yitao Liang and Song-Chun Zhu and Siyuan Huang},
  journal={ArXiv},
  year={2022},
  volume={abs/2210.07474}
}

@inproceedings{yang2025thinking,
  title={Thinking in space: How multimodal large language models see, remember, and recall spaces},
  author={Yang, Jihan and Yang, Shusheng and Gupta, Anjali W and Han, Rilyn and Fei-Fei, Li and Xie, Saining},
  booktitle=CVPR,
  pages={10632--10643},
  year={2025}
}

@article{Shen2023FlanMoESI,
  title={Flan-MoE: Scaling Instruction-Finetuned Language Models with Sparse Mixture of Experts},
  author={Sheng Shen and Le Hou and Yan-Quan Zhou and Nan Du and S. Longpre and Jason Wei and Hyung Won Chung and Barret Zoph and William Fedus and Xinyun Chen and Tu Vu and Yuexin Wu and Wuyang Chen and Albert Webson and Yunxuan Li and Vincent Y. Zhao and Hongkun Yu and Kurt Keutzer and Trevor Darrell and Denny Zhou},
  journal={ArXiv},
  year={2023},
  volume={abs/2305.14705}
}

@article{jiang2024mixtral,
  title={Mixtral of experts},
  author={Jiang, Albert Q and Sablayrolles, Alexandre and Roux, Antoine and Mensch, Arthur and Savary, Blanche and Bamford, Chris and Chaplot, Devendra Singh and Casas, Diego de las and Hanna, Emma Bou and Bressand, Florian and others},
  journal={ArXiv},
  year={2024},
  volume={abs/2401.04088}
}

@inproceedings{xue2024openmoe,
  title={OpenMoE: an early effort on open mixture-of-experts language models},
  author={Xue, Fuzhao and Zheng, Zian and Fu, Yao and Ni, Jinjie and Zheng, Zangwei and Zhou, Wangchunshu and You, Yang},
  booktitle=ICML,
  pages={55625--55655},
  year={2024}
}

@article{lin2026moe,
  title={MoE-LLaVA: Mixture of Experts for Large Vision-Language Models},
  author={Bin Lin and Zhenyu Tang and Yang Ye and Jiaxi Cui and Bin Zhu and Peng Jin and Jinfa Huang and Junwu Zhang and Munan Ning and Li Yuan},
  journal={ArXiv},
  year={2024},
  volume={abs/2401.15947}
}

@article{li2025uni,
  title={Uni-MoE: Scaling Unified Multimodal LLMs With Mixture of Experts},
  author={Yunxin Li and Shenyuan Jiang and Baotian Hu and Longyue Wang and Wanqi Zhong and Wenhan Luo and Lin Ma and Min Zhang},
  journal=PAMI,
  year={2024},
  volume={47},
  pages={3424-3439}
}

@inproceedings{zhang2024lmmseval,
  title={LMMs-Eval: Reality Check on the Evaluation of Large Multimodal Models},
  author={Kaichen Zhang and Bo Li and Peiyuan Zhang and Fanyi Pu and Joshua Adrian Cahyono and Kairui Hu and Yuhao Dong and Shuai Liu and Yuanhan Zhang and Jingkang Yang and Chunyuan Li and Ziwei Liu},
  booktitle={Findings of the Association for Computational Linguistics},
  volume={{NAACL} 2025},
  pages={881--916},
  year={2025}
}

@inproceedings{wang2025vggt,
  title={VGGT: Visual Geometry Grounded Transformer},
  author={Wang, Jianyuan and Chen, Minghao and Karaev, Nikita and Vedaldi, Andrea and Rupprecht, Christian and Novotny, David},
  booktitle=CVPR,
  pages={5294--5306},
  year={2025},
  doi={10.1109/CVPR52734.2025.00499}
}

@article{yang2024qwen2technicalreport,
      title={Qwen2 Technical Report}, 
      author={An Yang and Baosong Yang and Binyuan Hui and Bo Zheng and Bowen Yu and Chang Zhou and others},
      journal={ArXiv},
      year={2024},
      volume={abs/2407.10671} 
}

@article{qwen2025qwen25technicalreport,
      title={Qwen2.5 Technical Report}, 
      author={An Yang and Baosong Yang and Beichen Zhang and Binyuan Hui and Bo Zheng and Bowen Yu and Chengyuan Li and Dayiheng Liu and Fei Huang and others},
      journal={ArXiv},
      year={2024},
      volume={abs/2412.15115} 
}

@misc{yang2025qwen3technicalreport,
      title={Qwen3 Technical Report}, 
      author={An Yang and Anfeng Li and Baosong Yang and Beichen Zhang and Binyuan Hui and Bo Zheng and others},
      year={2025},
      eprint={2505.09388},
      archivePrefix={arXiv},
      primaryClass={cs.CL},
      url={https://arxiv.org/abs/2505.09388}, 
}

@inproceedings{Dai2024DeepSeekMoETU,
  title={DeepSeekMoE: Towards Ultimate Expert Specialization in Mixture-of-Experts Language Models},
  author={Damai Dai and Chengqi Deng and Chenggang Zhao and Runxin Xu and Huazuo Gao and Deli Chen and Jiashi Li and Wangding Zeng and Xingkai Yu and Yu Wu and Zhenda Xie and Y. K. Li and Panpan Huang and Fuli Luo and Chong Ruan and Zhifang Sui and Wenfeng Liang},
  booktitle={Proceedings of the 62nd Annual Meeting of the Association for Computational Linguistics},
  volume={1},
  pages={1280--1297},
  year={2024},
  doi={10.18653/V1/2024.ACL-LONG.70}
}

@inproceedings{azuma2022scanqa,
  title={Scanqa: 3d question answering for spatial scene understanding},
  author={Azuma, Daichi and Miyanishi, Taiki and Kurita, Shuhei and Kawanabe, Motoaki},
  booktitle=CVPR,
  pages={19129--19139},
  year={2022}
}

@misc{openai2024gpt4ocard,
      title={GPT-4o System Card}, 
      author={Aaron Hurst and Adam Lerer and Adam P. Goucher and Adam Perelman and Aditya Ramesh and Aidan Clark and AJ Ostrow and Akila Welihinda and Alan Hayes and Alec Radford and Aleksander Mądry and Alex Baker-Whitcomb and et al.},
      year={2024},
      eprint={2410.21276},
      archivePrefix={arXiv},
      primaryClass={cs.CL},
      url={https://arxiv.org/abs/2410.21276}, 
}

@article{geminiteam2024gemini15unlockingmultimodal,
  title={Gemini 1.5: Unlocking multimodal understanding across millions of tokens of context},
  author={Machel Reid and Nikolay Savinov and Denis Teplyashin and Dmitry Lepikhin and others},
  journal={ArXiv},
  year={2024},
  volume={abs/2403.05530}
}

@inproceedings{chen2024internvl,
  title={Internvl: Scaling up vision foundation models and aligning for generic visual-linguistic tasks},
  author={Chen, Zhe and Wu, Jiannan and Wang, Wenhai and Su, Weijie and Chen, Guo and Xing, Sen and Zhong, Muyan and Zhang, Qinglong and Zhu, Xizhou and Lu, Lewei and others},
  booktitle=CVPR,
  pages={24185--24198},
  year={2024}
}

@misc{zhu2025internvl3exploringadvancedtraining,
      title={InternVL3: Exploring Advanced Training and Test-Time Recipes for Open-Source Multimodal Models}, 
      author={Jinguo Zhu and Weiyun Wang and Zhe Chen and Zhaoyang Liu and Shenglong Ye and Lixin Gu and Hao Tian and Yuchen Duan and Weijie Su and others},
      year={2025},
      eprint={2504.10479},
      archivePrefix={arXiv},
      primaryClass={cs.CV},
      url={https://arxiv.org/abs/2504.10479}, 
}

@misc{chen2024longvilascalinglongcontextvisual,
      title={LongVILA: Scaling Long-Context Visual Language Models for Long Videos}, 
      author={Yukang Chen and Fuzhao Xue and Dacheng Li and Qinghao Hu and Ligeng Zhu and Xiuyu Li and Yunhao Fang and Haotian Tang and Shang Yang and Zhijian Liu and Ethan He and Hongxu Yin and Pavlo Molchanov and Jan Kautz and Linxi Fan and Yuke Zhu and Yao Lu and Song Han},
      year={2024},
      eprint={2408.10188},
      archivePrefix={arXiv},
      primaryClass={cs.CV},
      url={https://arxiv.org/abs/2408.10188}, 
}

@inproceedings{lin2024vila,
  title={Vila: On pre-training for visual language models},
  author={Lin, Ji and Yin, Hongxu and Ping, Wei and Molchanov, Pavlo and Shoeybi, Mohammad and Han, Song},
  booktitle=CVPR,
  pages={26689--26699},
  year={2024}
}

@article{Zhang2024LongCT,
  title={Long Context Transfer from Language to Vision},
  author={Peiyuan Zhang and Kaichen Zhang and Bo Li and Guangtao Zeng and Jingkang Yang and Yuanhan Zhang and Ziyue Wang and Haoran Tan and Chunyuan Li and Ziwei Liu},
  journal=TMLR,
  year={2025},
  volume={2025}
}

@misc{zhang2024llavanextvideo,
  title={LLaVA-NeXT: A Strong Zero-shot Video Understanding Model},
  url={https://llava-vl.github.io/blog/2024-04-30-llava-next-video/},
  author={Zhang, Yuanhan and Li, Bo and Liu, haotian and Lee, Yong jae and Gui, Liangke and Fu, Di and Feng, Jiashi and Liu, Ziwei and Li, Chunyuan},
  month={April},
  year={2024}
}

@misc{li2024llavaonevisioneasyvisualtask,
      title={LLaVA-OneVision: Easy Visual Task Transfer}, 
      author={Bo Li and Yuanhan Zhang and Dong Guo and Renrui Zhang and Feng Li and Hao Zhang and Kaichen Zhang and Peiyuan Zhang and Yanwei Li and Ziwei Liu and Chunyuan Li},
      year={2024},
      eprint={2408.03326},
      archivePrefix={arXiv},
      primaryClass={cs.CV},
      url={https://arxiv.org/abs/2408.03326}, 
}

@misc{ouyang2025spacerreinforcingmllmsvideo,
      title={SpaceR: Reinforcing MLLMs in Video Spatial Reasoning}, 
      author={Kun Ouyang and Yuanxin Liu and Haoning Wu and Yi Liu and Hao Zhou and Jie Zhou and Fandong Meng and Xu Sun},
      year={2025},
      eprint={2504.01805},
      archivePrefix={arXiv},
      primaryClass={cs.CV},
      url={https://arxiv.org/abs/2504.01805}, 
}

@misc{wu2025reinforcingspatialreasoningvisionlanguage,
      title={Reinforcing Spatial Reasoning in Vision-Language Models with Interwoven Thinking and Visual Drawing}, 
      author={Junfei Wu and Jian Guan and Kaituo Feng and Qiang Liu and Shu Wu and Liang Wang and Wei Wu and Tieniu Tan},
      year={2025},
      eprint={2506.09965},
      archivePrefix={arXiv},
      primaryClass={cs.CV},
      url={https://arxiv.org/abs/2506.09965}, 
}

@inproceedings{zhu20233dvista,
  title={3d-vista: Pre-trained transformer for 3d vision and text alignment},
  author={Zhu, Ziyu and Ma, Xiaojian and Chen, Yixin and Deng, Zhidong and Huang, Siyuan and Li, Qing},
  booktitle=ICCV,
  pages={2911--2921},
  year={2023}
}

@inproceedings{chen2024ll3da,
  title={Ll3da: Visual interactive instruction tuning for omni-3d understanding reasoning and planning},
  author={Chen, Sijin and Chen, Xin and Zhang, Chi and Li, Mingsheng and Yu, Gang and Fei, Hao and Zhu, Hongyuan and Fan, Jiayuan and Chen, Tao},
  booktitle=CVPR,
  pages={26428--26438},
  address = {Seattle, WA, USA},
  year={2024}
}

@inproceedings{deng20253d,
  title={3d-llava: Towards generalist 3d lmms with omni superpoint transformer},
  author={Deng, Jiajun and He, Tianyu and Jiang, Li and Wang, Tianyu and Dayoub, Feras and Reid, Ian},
  booktitle=CVPR,
  pages={3772--3782},
  year={2025}
}

@inproceedings{zheng2025video,
  title={Video-3d llm: Learning position-aware video representation for 3d scene understanding},
  author={Zheng, Duo and Huang, Shijia and Wang, Liwei},
  booktitle=CVPR,
  pages={8995--9006},
  year={2025}
}

@misc{liu2025oryxmllmondemandspatialtemporal,
      title={Oryx MLLM: On-Demand Spatial-Temporal Understanding at Arbitrary Resolution}, 
      author={Zuyan Liu and Yuhao Dong and Ziwei Liu and Winston Hu and Jiwen Lu and Yongming Rao},
      year={2025},
      eprint={2409.12961},
      archivePrefix={arXiv},
      primaryClass={cs.CV},
      url={https://arxiv.org/abs/2409.12961}, 
}

@article{zhu2025struct2d,
  title={Struct2D: A Perception-Guided Framework for Spatial Reasoning in MLLMs},
  author={Zhu, Fangrui and Wang, Hanhui and Xie, Yiming and Gu, Jing and Ding, Tianye and Yang, Jianwei and Jiang, Huaizu},
  journal={arXiv preprint arXiv:2506.04220},
  year={2025}
}

\clearpage


\section*{Supplementary material}

# SpaR3D-MoE: Adaptive 3D Spatial Reasoning from Sparse Views Meets Geometry-Inductive Mixture-of-Experts

## Supplementary Material

Haida Feng[1,2], Hao Wei[1,3†], Haolin Wang[1,2], Shiwei Li[1,2], Chade Li[1,2], and Yihong Wu[1,2,3†]

[1] State Key Laboratory of Multimodal Artificial Intelligence Systems, Institute of Automation, Chinese Academy of Sciences, China
[2] School of Artificial Intelligence, University of Chinese Academy of Sciences, China
[3] YUKUN Intelligent World, Beijing, China
{fenghaida2024@ia, weihao2019@ia, wanghaolin2023@ia, lishiwei2023@ia, lichade2021@ia, yhwu@nlpr.ia}.ac.cn

# 1 Additional Experimental Details

## 1.1 Implementation Details

SpaR3D-MoE is built upon the pre-trained Qwen3VL-8B [2] and incorporates VGGT-1B [9] as the visual geometry encoder. During training, both the visual and geometry encoders are frozen, while we optimize the HGI-MoE, multimodal projector, and the MLLM backbone. Specifically, the MLLM backbone is fine-tuned using LoRA [6] with rank $r = 256$ and scaling factor $\alpha = 512$. The network is optimized using AdamW [7] with a weight decay of 0.01. We apply a cosine learning rate scheduler with a 3% warmup ratio, setting the peak learning rate to $5 \times 10^{-5}$ for the MLLM backbone and $2 \times 10^{-4}$ for the remaining trainable modules, alongside a global batch size of 128. For the dynamic routing within the HGI-MoE, we configure the Top-$K$ gate to activate 3 experts ($K = 3$) for each multimodal token. To ensure balanced expert routing within the MoE module, we apply an auxiliary load-balancing loss with a coefficient of 0.01. The training inputs consist of non-uniformly sampled sparse keyframes, driving the model to capture and reason over varying spatiotemporal dynamics. For training data, we utilize a mixed dataset consisting of a 288K subset from VSI-590K [11], alongside the training sets of VICA-322K [5], ScanQA [1], and SQA3D [8]. The entire training process runs for one epoch and takes approximately 11 hours on 5 $\times$ NVIDIA RTX PRO 6000 Blackwell (96GB) GPUs.

† Corresponding authors

### 1.2 Evaluation Datasets and Metrics

We utilize the LMMs-Eval framework [13] for standardized evaluation across all benchmarks, extending it with custom evaluation scripts to seamlessly integrate unsupported datasets and ensure unified metric computation. To guarantee deterministic and reproducible results, we employ greedy decoding (temperature $\tau = 0$, beam size 1). For our primary baseline, Qwen3VL, we conduct local evaluations under identical settings, while results for other methods are directly quoted from their original publications to ensure consistency and fairness in comparison. The specific evaluation datasets and metrics are introduced as follows.

**VSI-Bench Dataset.** VSI-Bench [10] is a comprehensive video-based benchmark designed to evaluate the visual-spatial intelligence of MLLMs. It contains over 5,000 QA pairs derived from 288 real-world indoor egocentric videos sourced from 3D environments like ScanNet [4], ScanNet++ [12], and ARKitScenes [3]. These QA pairs are structured to evaluate 8 fine-grained spatial reasoning tasks, which are categorized into two distinct formats: Multiple-Choice Answer (MCA) tasks and Numerical Answer (NA) tasks. For evaluation metrics, we follow the official protocol, adopting standard Accuracy for MCA tasks and Mean Relative Accuracy (MRA) for NA tasks.

**ScanQA Dataset.** ScanQA [1] is a foundational 3D visual question answering benchmark designed to assess a model's spatial reasoning and semantic grounding capabilities within complex indoor environments. We conduct our evaluation using the official validation split, which consists of over 4,000 question-answering pairs derived from real-world scenes in the ScanNet dataset [4]. Consistent with the official evaluation protocol, we report Exact Match (EM) accuracy to measure strict answer correctness, and employ standard natural language generation metrics (BLEU-1 to BLEU-4, METEOR, ROUGE-L, and CIDEr) to evaluate the fluency and descriptive quality of the generated responses.

**SQA3D Dataset.** SQA3D [8] is a benchmark dedicated to situated question answering, requiring models to perceive and reason about 3D environments from a specific agent's egocentric position and orientation. We conduct our evaluation on the official test split, comprising over 3,000 question-answering pairs that are classified into six distinct reasoning categories (What, Is, How, Can, Which, and Other). Following the official protocol, we report the Exact Match at 1 (EM@1) accuracy to assess the performance within each individual question category, as well as the overall average score.

## 2 Additional Ablation Study

### 2.1 Component Decoupling Analysis

To disentangle the contributions of ASMS and HGI-MoE, we conduct a $2 \times 2$ cross-ablation on VSI-Bench by varying the sampling strategy and the inte-

**Table 6:** Cross-ablation of ASMS and HGI-MoE on VSI-Bench. All variants use 32 input frames.

| Sampling | Integration Architecture | Avg. Score | Gain |
|---|---|---|---|
| Uniform | Monolithic Fusion | 58.4 | – |
| ASMS | Monolithic Fusion | 59.3 | +0.9 |
| Uniform | HGI-MoE | 62.4 | +4.0 |
| ASMS | HGI-MoE | 63.5 | +5.1 |

**Table 7:** Robustness to upstream geometry and pose perturbations on VSI-Bench.

| Noise Type | Geo. $\sigma^2$ | Pose $\sigma^2$ | Avg. | $\Delta$ |
|---|---|---|---|---|
| Baseline (None) | 0.00 | 0.00 | 63.5 | – |
| Geo. Noise Only | 0.10 | 0.00 | 62.6 | -0.9 |
| Pose Noise Only | 0.00 | 0.10 | 61.8 | -1.7 |
| Combined Noise | 0.10 | 0.10 | 62.2 | -1.3 |

gration architecture. Specifically, we compare uniform sampling and ASMS under both monolithic visual-geometric fusion and the proposed HGI-MoE. As shown in Tab. 6, replacing uniform sampling with ASMS improves the sparse keyframe representation by preserving more informative spatiotemporal connectivity, while replacing monolithic fusion with HGI-MoE improves the integration architecture for adaptive multimodal fusion. Combining both components achieves the best performance, suggesting that topology-aware sparse sampling and adaptive heterogeneous integration provide complementary benefits for 3D spatial reasoning.

### 2.2 Robustness to Geometry and Pose Perturbations

Since SpaR3D-MoE relies on VGGT-predicted geometry features and camera poses for sampling and fusion, we evaluate its sensitivity to upstream prediction noise. Specifically, we inject Gaussian noise with variance $\sigma^2 = 0.1$ into the normalized geometry features and/or pose features extracted by VGGT, and report the results on VSI-Bench in Tab. 7. The model shows a moderate performance drop under both geometry and pose perturbations, indicating that accurate upstream geometry is beneficial for spatial reasoning. Nevertheless, SpaR3D-MoE maintains competitive performance under all perturbation settings, suggesting a certain degree of robustness to imperfect geometry and pose estimates.

**Table 8:** Ablation on the temporal coefficient $\omega$ in ASMS on VSI-Bench. We report average scores for the Numerical and Multiple-Choice Answer categories. All experiments are evaluated via LMMs-Eval [13] using 32 frames.

| **Method** | | **Avg.** | **Numerical Answer** | **Multiple-Choice Answer** |
|---|---|---|---|---|
| Qwen3VL-8B (Baseline) | | 55.7 | **61.9** | 49.5 |
| + ASMS | $\omega = 0.1$ | 55.1 | 60.9 | 49.3 |
| | $\omega = 0.6$ (Default) | **56.2** | 61.3 | **51.2** |
| | $\omega = 1.1$ | 55.6 | 60.9 | 50.3 |

### 2.3 Hyperparameter Sensitivity in ASMS

To analyze the temporal coefficient $\omega$ in ASMS, we extract 32 sparse frames under varying $\omega$ values and evaluate them on VSI-Bench using Qwen3VL-8B, comparing against a 32-frame uniform sampling baseline. This hyperparameter controls the balance between geometric diversity and temporal stability. As shown in Tab. 8, the model performance exhibits a clear trade-off, peaking at the intermediate value. Specifically, setting a low $\omega = 0.1$ overly focuses on spatial differences to extract visually diverse frames, but disrupts the temporal progression, reducing accuracy on Multiple-Choice Answer (MCA) tasks (49.3) that require trajectory reasoning. Conversely, setting a high $\omega = 1.1$ makes the temporal penalty dominate, reducing ASMS to a uniform sampling that misses crucial spatial transitions and causes topological disconnection, yielding a average score of 55.6. Notably, both extreme configurations underperform the uniform sampling baseline (55.7), demonstrating that improper temporal constraints degrade spatial reasoning. Ultimately, setting $\omega = 0.6$ achieves the optimal balance. By preserving the spatiotemporal manifold while capturing essential geometric diversity, this setting maximizes the informativeness of the sparse keyframes, resulting in the highest overall average (56.2) and peak MCA accuracy (51.2).

### 2.4 Efficiency Analysis

We report the system-level efficiency of SpaR3D-MoE on VSI-Bench in Tab. 9. The 3D encoding stage includes VGGT-based feature and pose extraction as well as ASMS keyframe selection, where VGGT extraction takes 7.88s per video and ASMS adds only 0.13s overhead. This scene-level preprocessing makes the initial query take 12.54s, comparable to Spatial-MLLM-4B (12.69s). Within the reasoning stage, the router and MoE introduce only minor overheads of approximately 0.06s and 0.02s, respectively, while SpaR3D-MoE activates about 9B out of 10B total parameters under Top-$K = 3$ routing. Importantly, the preprocessing cost is mainly incurred only when processing a new scene for the first time. For an unchanged scene, the extracted geometry features, camera poses,

**Table 9:** Latency and memory efficiency on VSI-Bench. The 3D encoding time of SpaR3D-MoE includes VGGT feature extraction and ASMS keyframe selection. Subsequent in-scene question answering can reuse cached scene representations and only requires the reasoning stage.

| Model | 2D Encoding Time (s) | 3D Encoding Time (s) | Reasoning Time (s) | Initial Query Time (s) | Peak Memory (GB) |
|---|---|---|---|---|---|
| Spatial-MLLM-4B | 0.36 | 8.93 | 3.40 | 12.69 | 13.10 |
| Qwen3VL-8B | 0.72 | N/A | 2.37 | 3.09 | 17.91 |
| SpaR3D-MoE | 0.39 | 8.01 | 4.14 | 12.54 | 27.70 |

and sparse keyframes can be cached and reused, so subsequent in-scene question answering only requires the reasoning stage and takes approximately 4.14s.

# 3 Qualitative Analysis

## 3.1 Qualitative Comparison of Sampling Strategies

We qualitatively compare our ASMS mechanism with the uniform sampling baseline under a sparse 8-frame setting in Fig. 4. For an intuitive comparison, we utilize VGGT [9] to explicitly project the extracted implicit geometric features into a global point cloud, where the corresponding camera poses are marked. As illustrated in Fig. 4(a), uniform sampling extracts frames at fixed temporal intervals, resulting in the omission of critical spatial transitions due to its rigid, content-agnostic nature. Specifically, this topology-agnostic approach fails to capture the doorway transition (indicated by dashed orange boxes) resulting in topological disconnection and a fragmented global point cloud, which compromises global spatial reasoning.

In contrast, Fig. 4(b) demonstrates that ASMS adaptively captures crucial bridge frames (solid green box) by jointly evaluating 3D spatial translations, geometric variance, and temporal dynamics. Preserving these transitional nodes ensures that the extracted sparse frames maintain the underlying topological connectivity of the original scene. Consequently, ASMS is able to provide high-quality foundational scene context, supporting complex downstream spatial tasks, such as relative distance estimation and route planning.

## 3.2 Qualitative Analysis on 3D Spatial Reasoning

**Analysis of Benchmark Scenarios.** To further demonstrate the effectiveness of SpaR3D-MoE, we present qualitative comparisons against the Qwen3VL-8B baseline on two distinct spatial reasoning tasks from the VSI-Bench [10]. As illustrated in Fig. 5, we present a Relative Direction task where the agent is required to reason about the spatial relationship between a trash bin and its own location. The baseline Qwen3VL fails to build a consistent 3D representation from sparse 2D views, leading to an incorrect "back-right" prediction. In contrast,

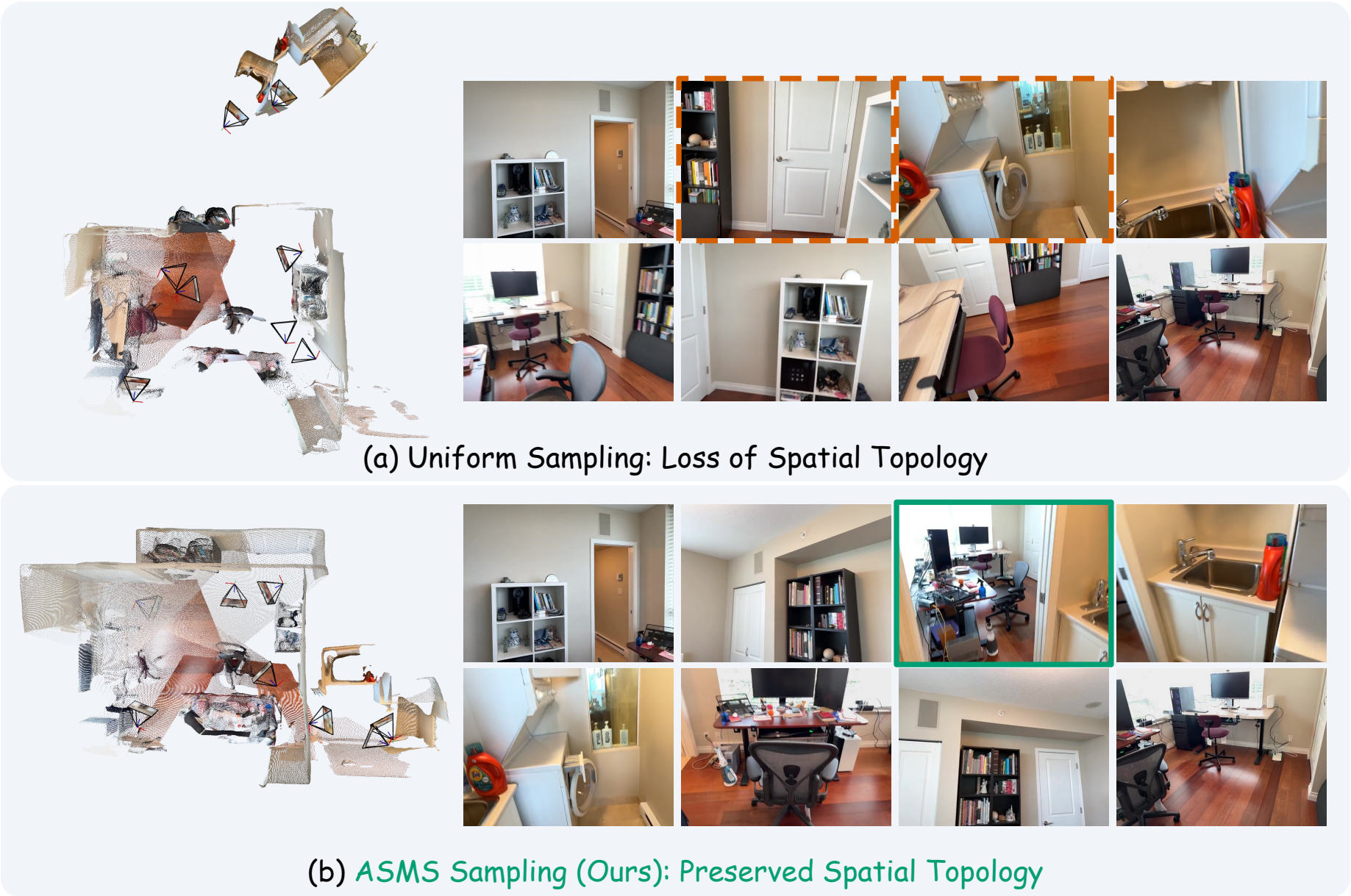


**Fig. 4: Qualitative comparison between ASMS and uniform sampling under a sparse 8-frame setting. (a)** Uniform sampling misses key spatial transitions (dashed orange boxes), resulting in topological disconnection and a fragmented point cloud. **(b)** Our ASMS adaptively captures crucial bridge frames (solid green box) to preserve the global spatial topology, yielding a coherent scene structure.

our model successfully grounds these 2D observations into a unified 3D scene, enabling it to correctly reason the "front-left" spatial relationship. Furthermore, Fig. 6 showcases a Route Planning task requiring egocentric navigation. The baseline Qwen3VL incorrectly predicts "Turn Right" due to misleading 2D visual information, where the target doorframe appears on the right side of the input image frame. Conversely, SpaR3D-MoE correctly reasons that the doorframe is located behind the agent's current orientation, yielding the accurate "Turn Back" action. This spatial reasoning capability stems from our ASMS, which provides rich scene context while preserving topological connectivity from sparse RGB views, coupled with the HGI-MoE architecture that dynamically activates specialized experts to fuse multimodal geometric and pose features, enabling the agent to construct a unified 3D spatial representation and capture egocentric motion dynamics. By grounding sparse visual inputs into a unified physical space, SpaR3D-MoE overcomes inherent 2D projection ambiguities to perform accurate 3D spatial reasoning.

**Generalization to Real-World Scenes.** To further validate the zero-shot generalization of SpaR3D-MoE, we conduct qualitative tests on a custom-captured real-world indoor lab scene. As illustrated in Fig. 7, this task requires the model

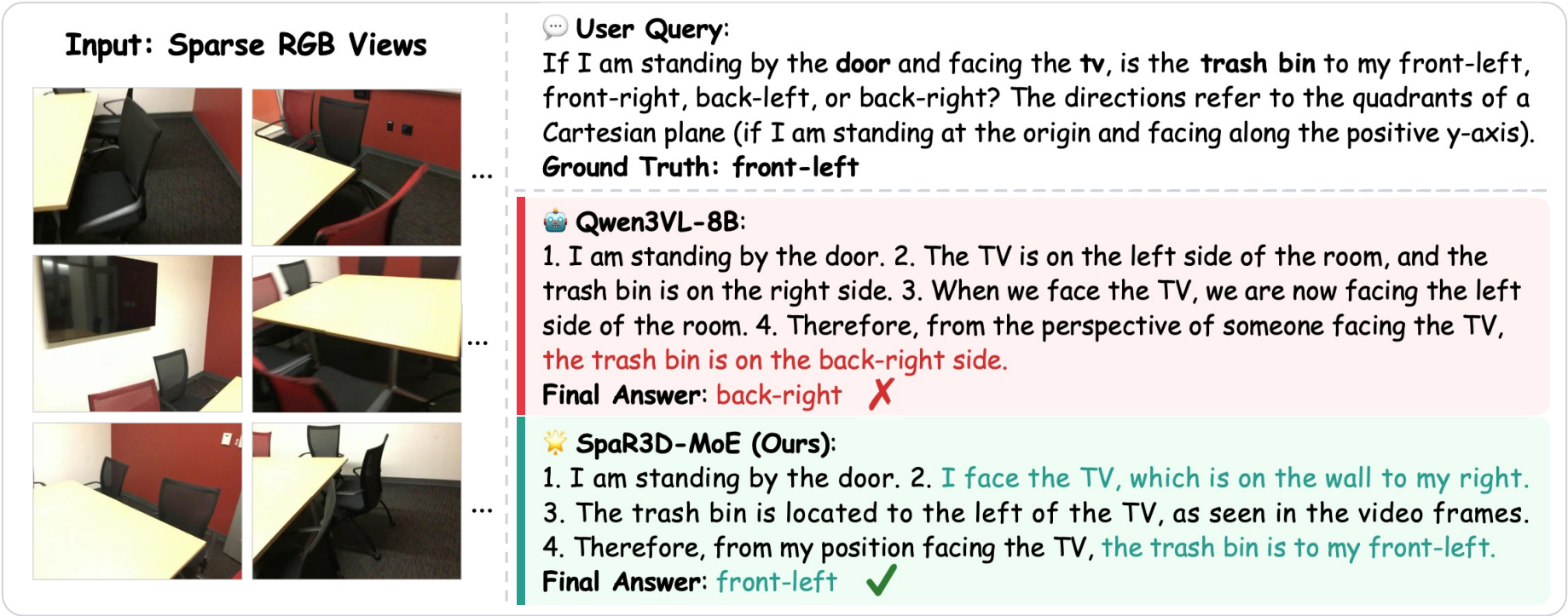


**Fig. 5: Qualitative Results on Relative Direction.** The baseline Qwen3VL fails to resolve the viewpoint-dependent transformation, leading to spatial hallucinations based on 2D image coordinates. In contrast, SpaR3D-MoE successfully bridges the gap between 2D semantics and 3D geometry, accurately grounding the target's position within the agent's egocentric perspective.

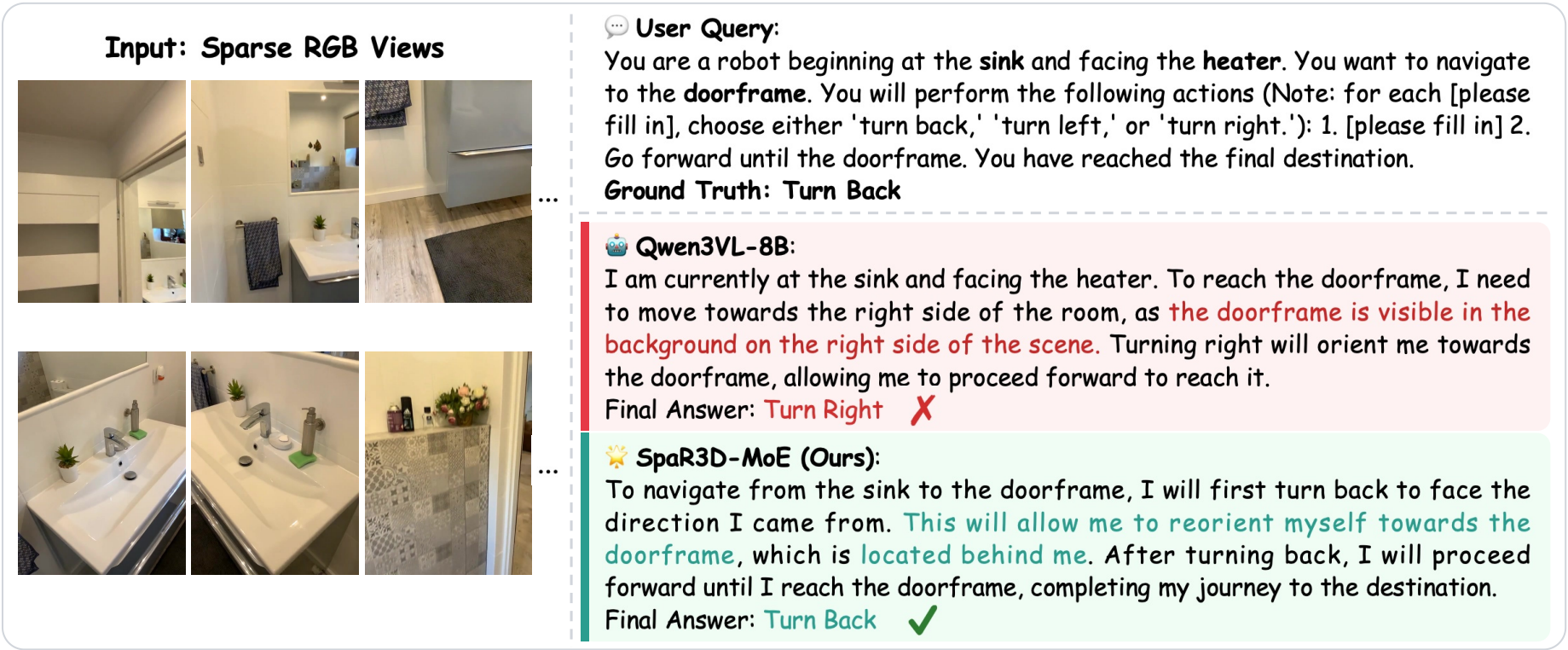


**Fig. 6: Qualitative Results on Route Planning.** The baseline is biased by the 2D visual information where the target doorframe appears on the right side of the image frame, leading to an incorrect prediction. Conversely, SpaR3D-MoE leverages a consistent egocentric spatial map to correctly reason the agent's egocentric orientation, yielding the accurate "Turn Back" action.

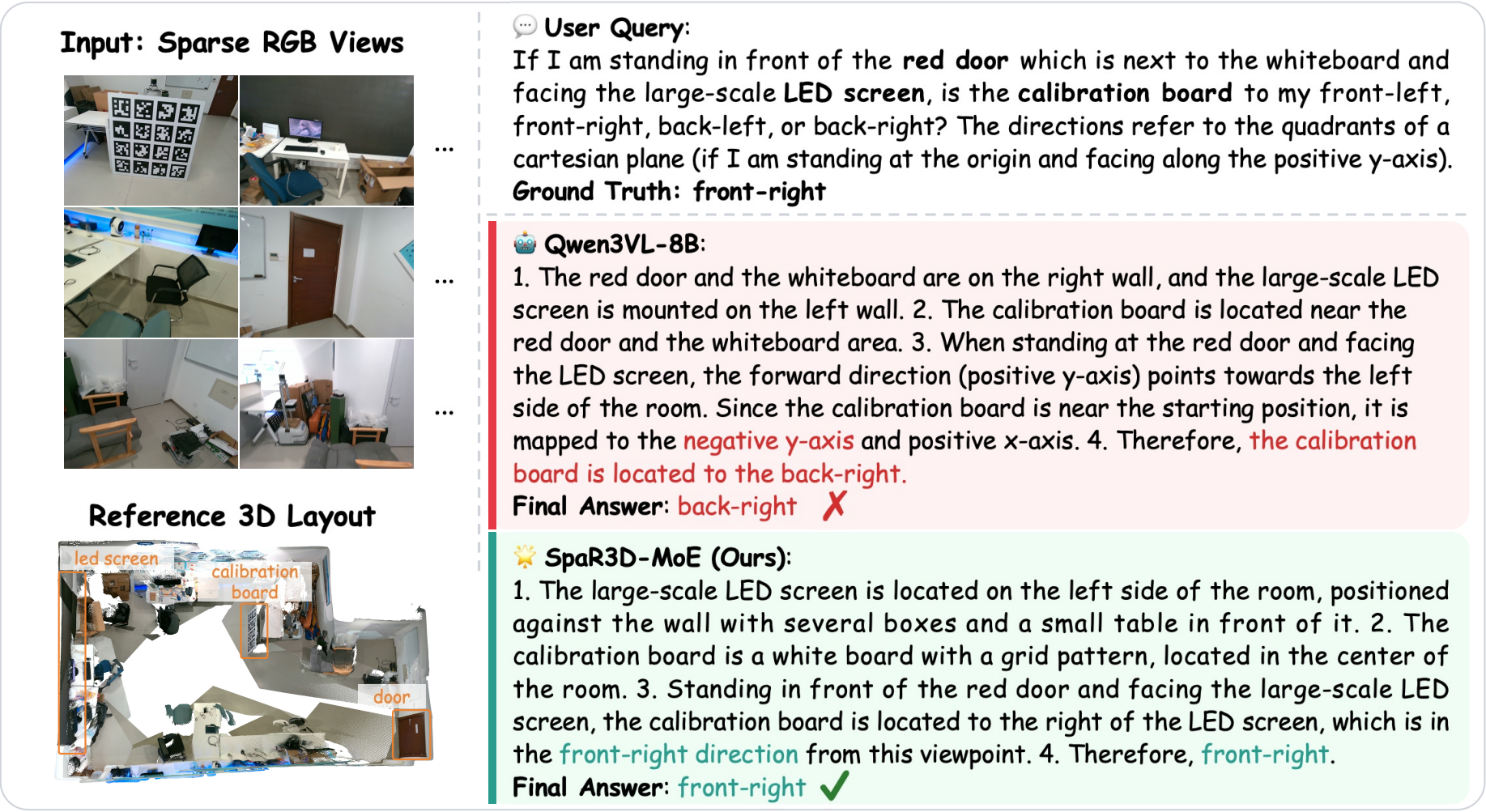


**Fig. 7: Qualitative results in an unseen laboratory scene.** In an unseen laboratory setting, the baseline incorrectly maps sparse observations to the spatial coordinate system, localizing the target along the negative $y$-axis. In contrast, SpaR3D-MoE accurately grounds visual cues within a unified physical reference frame, enabling precise reasoning of relative positions from sparse views.

to perform coordinate-based relative direction reasoning within a real-world laboratory scene using only sparse views. The baseline Qwen3VL incorrectly maps sparse 2D visual observations to the spatial coordinate system, leading to a negative y-axis direction and an incorrect "back-right" answer. In contrast, by grounding the red door, LED screen, and calibration board within a unified physical reference frame, our SpaR3D-MoE correctly reasons the "front-right" relative position. These results demonstrate that our method successfully generalizes to real-world scenes using only sparse RGB views, effectively bridging the representational gap between 2D visual semantics and 3D physical geometry to enable robust spatial reasoning.

### 3.3 Limitations and Failure Analysis

While SpaR3D-MoE demonstrates robust spatial reasoning and generalization capabilities in most scenarios, it still faces challenges with precise instance disambiguation in environments containing repetitive objects. As illustrated in Fig. 8, the model is required to reason about the relative distances to identify the object nearest to the backpack. The thinking step reveals that the model successfully localizes all candidates and correctly associates them with their respective semantic surfaces (e.g., the backpack and refrigerator on the floor, the pillow on the bed). However, the model exhibits a limitation in handling instance ambiguity during distance comparison. Specifically, while the distance to the refrigerator is reasonably estimated (∼1.5m), the distance to the pillow is overestimated (∼3.5m)

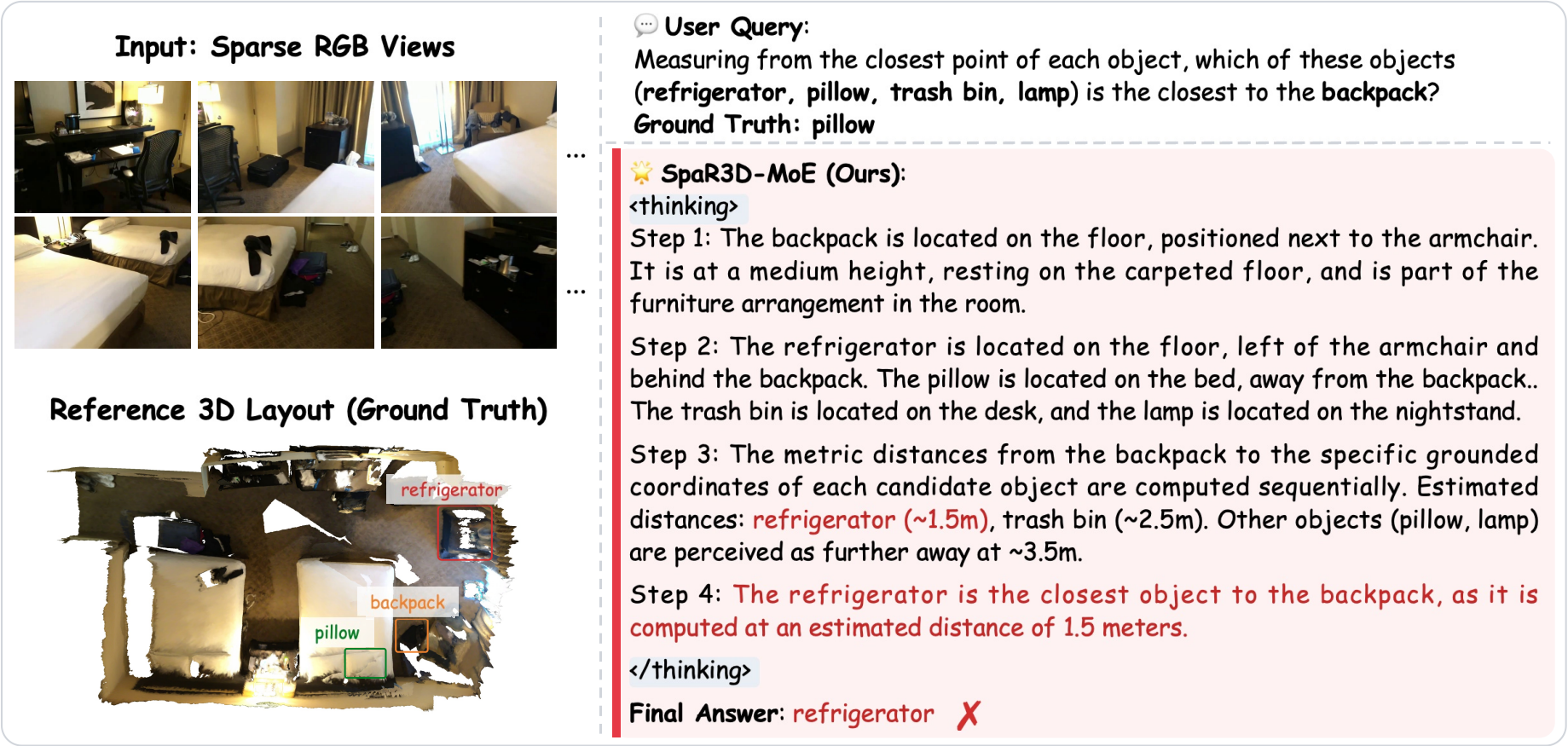


**Fig. 8: Qualitative failure case of instance ambiguity.** The green box denotes the ground-truth target (pillow), the orange box indicates the reference anchor (backpack), and the red box represents the model's incorrect final prediction (refrigerator). While SpaR3D-MoE correctly identifies semantic categories, it misgrounds the target "pillow" to a distant instance on the opposite bed. Consequently, the distance between the pillow and the backpack is erroneously estimated at ∼3.5m, leading to the incorrect selection of the refrigerator.

because the model incorrectly anchors the semantic token to a distant pillow on the opposite bed. Consequently, this leads to the incorrect final prediction that the refrigerator is the closest object to the backpack. Future improvements will explore integrating Reinforcement Learning (RL) driven by geometric feedback to enhance the model's self-verification capabilities. Cross-validating candidate instances allows the model to calibrate its spatial anchoring, thereby penalizing referential misgrounding and improving instance disambiguation.

## References


1. Azuma, D., Miyanishi, T., Kurita, S., Kawanabe, M.: Scanqa: 3d question answering for spatial scene understanding. In: IEEE Conf. Comput. Vis. Pattern Recog. pp. 19129–19139 (2022)
2. Bai, S., Cai, Y., Chen, R., Chen, K., et al.: Qwen3-vl technical report. ArXiv **abs/2511.21631** (2025)
3. Baruch, G., Chen, Z., Dehghan, A., Dimry, T., Feigin, Y., Fu, P., Gebauer, T., Joffe, B., Kurz, D., Schwartz, A., Shulman, E.: ARKitscenes - a diverse real-world dataset for 3d indoor scene understanding using mobile RGB-d data. In: Thirty-fifth Conference on Neural Information Processing Systems Datasets and Benchmarks Track (Round 1) (2021)
4. Dai, A., Chang, A.X., Savva, M., Halber, M., Funkhouser, T., Nießner, M.: Scannet: Richly-annotated 3d reconstructions of indoor scenes. In: Proceedings of the IEEE conference on computer vision and pattern recognition. pp. 5828–5839 (2017)
5. Feng, Q.: Visuospatial cognitive assistant. ArXiv **abs/2505.12312** (2025)
6. Hu, E.J., Shen, Y., Wallis, P., Allen-Zhu, Z., Li, Y., Wang, S., Wang, L., Chen, W.: LoRA: Low-rank adaptation of large language models. In: International Conference on Learning Representations (2022)
7. Loshchilov, I., Hutter, F.: Decoupled weight decay regularization. In: International Conference on Learning Representations (2017)
8. Ma, X., Yong, S., Zheng, Z., Li, Q., Liang, Y., Zhu, S.C., Huang, S.: Sqa3d: Situated question answering in 3d scenes. ArXiv **abs/2210.07474** (2022)
9. Wang, J., Chen, M., Karaev, N., Vedaldi, A., Rupprecht, C., Novotny, D.: Vggt: Visual geometry grounded transformer. In: IEEE Conf. Comput. Vis. Pattern Recog. pp. 5294–5306 (2025). `https://doi.org/10.1109/CVPR52734.2025.00499`
10. Yang, J., Yang, S., Gupta, A.W., Han, R., Fei-Fei, L., Xie, S.: Thinking in space: How multimodal large language models see, remember, and recall spaces. In: IEEE Conf. Comput. Vis. Pattern Recog. pp. 10632–10643 (2025)
11. Yang, S., Yang, J., Huang, P., Brown, E., Yang, Z., Yu, Y., Tong, S., Zheng, Z., Xu, Y., Wang, M., Lu, D., Fergus, R., LeCun, Y., Li, F.F., Xie, S.: Cambrian-s: Towards spatial supersensing in video. ArXiv **abs/2511.04670** (2025)
12. Yeshwanth, C., Liu, Y.C., Nießner, M., Dai, A.: Scannet++: A high-fidelity dataset of 3d indoor scenes. In: Proceedings of the IEEE/CVF International Conference on Computer Vision. pp. 12–22 (2023)
13. Zhang, K., Li, B., Zhang, P., Pu, F., Cahyono, J.A., Hu, K., Dong, Y., Liu, S., Zhang, Y., Yang, J., Li, C., Liu, Z.: Lmms-eval: Reality check on the evaluation of large multimodal models. In: Findings of the Association for Computational Linguistics. vol. NAACL 2025, pp. 881–916 (2025)



\end{document}